\pdfoutput=1

\documentclass[11pt]{article}

\usepackage[final]{acl}

\usepackage{times}
\usepackage{latexsym}

\usepackage[T1]{fontenc}

\usepackage[utf8]{inputenc}

\usepackage{microtype}

\usepackage{inconsolata}

\usepackage{graphicx}

\usepackage[utf8]{inputenc} 
\usepackage[T1]{fontenc}    
\usepackage{hyperref}       
\usepackage{url}            
\usepackage{booktabs}       
\usepackage{amsfonts}       
\usepackage{nicefrac}       
\usepackage{xcolor}         
\usepackage{graphicx}
\usepackage[most]{tcolorbox}
\usepackage[table,xcdraw]{xcolor}
\usepackage{arydshln}
\usepackage{multirow} 
\usepackage{caption}
\usepackage{amsmath}
\usepackage{amssymb}
\usepackage{array}

\title{{\scshape ClinicalMC}: A Benchmark for Multi-Course Clinical Decision-Making with Large Language Models}

\author{
\textbf{Ruihui Hou\textsuperscript{\rm $\diamondsuit$}},
\textbf{Siyi Zhu\textsuperscript{\rm $\diamondsuit$}},
 \textbf{Ziyue Huai\textsuperscript{\rm $\diamondsuit$}},
 \textbf{Guangya Yu\textsuperscript{\rm $\diamondsuit$}},
 \textbf{Yongqi Fan\textsuperscript{\rm $\diamondsuit$}},
\\
 \textbf{Chunming Wang\textsuperscript{\rm $\clubsuit$}},
 \textbf{Tong Ruan\textsuperscript{\rm $\diamondsuit$}\thanks{~~Corresponding Authors.}}
\\
 \textsuperscript{\rm $\diamondsuit$}East China University of Science and Technology, Shanghai, China,
\\
 \textsuperscript{\rm $\clubsuit$}Renji Hospital Affiliated to Shanghai Jiaotong University\\
 School of Medicine, Shanghai, China.
 \\
}

\begin{document}
\maketitle
\begin{abstract}

Large language models (LLMs) have been widely adopted in healthcare, yet they still encounter significant challenges in complex clinical decision-making scenarios. 
Existing benchmarks primarily assess LLM performance in single-course settings and lack systematic evaluation in multi-course scenarios, where a patient’s condition evolves over time.
To address this gap, we propose ClinicalMC, a benchmark for multi-course clinical decision-making. 
It includes 1,275 Chinese and 5,804 English samples across four stages from admission to discharge. 
These stages cover triage, first-course examination/diagnosis/treatment, subsequent multi-course examination/assessment/treatment, and final diagnosis. 
In ClinicalMC, patients in the English dataset undergo an average of 5.11 clinical courses, whereas those in the Chinese dataset undergo 3.42.
To assess LLM performance, we construct a multi-agent evaluation framework that includes patient, examiner, and doctor agents.
Based on the benchmark and framework, we design two experimental settings—a single-turn static setting and a multi-turn dynamic setting—and assess three categories of LLMs:
1) closed-source LLMs like GPT5-mini; 2) open-source LLMs like DeepSeek-V3.2, and 3) medical LLMs like HuatuoGPT-o1. 
Through extensive evaluation, we aim to better understand LLM performance in the medical domain and support its effective deployment in healthcare.\footnote{Data and code are available at the URL \url{https://github.com/hzyuezh/ClinicalMPD}.}


\end{abstract}

\section{Introduction}
\label{introduction}

Large language models (LLMs) have shown strong performance in various medical NLP tasks, including information extraction~\cite{RAMIE}, text generation~\cite{lin2023survey} and question answering~\cite{medqa}. 
However, their reliability remains limited in complex clinical decision-making scenarios~\cite{MIMICCDM}, which require the continuous integration of heterogeneous data (e.g., vital signs, laboratory results) and real-time reasoning under evolving patient conditions~\cite{overview}.
This limitation highlights the necessity of systematically evaluating the LLM applications in multi-course\footnote{The course is a continuous record of a patient’s condition and treatment during hospitalization, including key details such as vital signs, surgeries, and major clinical changes.} clinical decision-making.


\begin{figure}[t]
\centering
\includegraphics[width=\columnwidth]{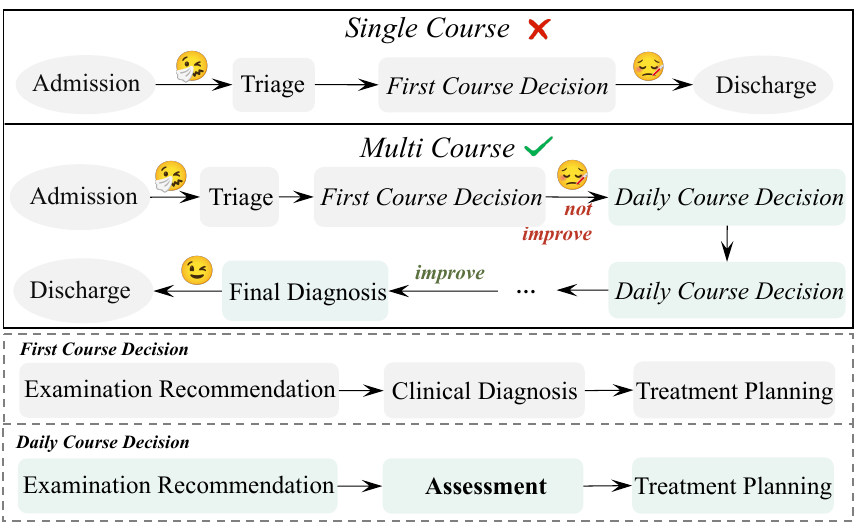} 
\caption{
The solid boxes highlight the distinctions between our clinical decision-making tasks and previous benchmarks. Both ``First Course Decision'' and ``Daily Course Decision'' consist of three subtasks each, while the dashed boxes provide their detailed descriptions.
}
\label{workflow}
\end{figure}

\begin{table*}[ht]
\small
\caption{Overview of clinical decision-making benchmarks. ``Dept.'', ``CAS.'', ``A2D.'', and ``Multi-C.'' stand for ``department'', ``assessment task'', ``admission to discharge process'', and ``multiple courses decision'', respectively. ``continuous assessment'' indicates whether the patient’s condition is continuously assessed.}
\resizebox{\textwidth}{!}{
\begin{tabular}{lccccccc}
\toprule
{\textbf{Dataset}}    &   \multicolumn{1}{c}{\textbf{Data Source}}   &   \multicolumn{1}{c}{\textbf{\# Dept.}}   &   \multicolumn{1}{c}{\textbf{A2D.}}  &   \multicolumn{1}{c}{\textbf{CAS.}}   &   \multicolumn{1}{c}{\textbf{Multi-C.}}     &     \multicolumn{1}{c}{\textbf{Language}}     &     \multicolumn{1}{c}{\textbf{\# EHRs}}     \\ \midrule
MedR-Bench~\cite{MedRBench}     & PMC-patients  & 10     & ×    & ×   & ×     & en    & 1,453   \\
ClinicalLab~\cite{clinicallab}  & Grade 3A Hospital   & 24    & \checkmark   & ×  & ×     & zh    & 1,500 \\
MIMIC-CDM~\cite{MIMICCDM}   & MIMIC-IV    & 1    & ×   & ×   & ×     & en    & 2,400     \\
MedChain~\cite{medchain}    & Medical Website    & 19     & \checkmark   & ×   & ×   & zh     & 12,163   \\
AI Hospital~\cite{AIHospital}  & Medical Website    & 6   & ×    & ×   & ×   & zh     & 506      \\
CRAFT-MD~\cite{CRAFTMD}   & MedQA    & 12     & ×   & ×    & ×    & en       & 2,000   \\
MAP~\cite{Map}    & MIMIC-IV     & 9     & ×    & ×    & ×    & en      & 51,274     \\
MedJourney~\cite{MedJourney}   & Smart Hospital   & 12  & ×    & ×   & ×    & zh   & 4,960     \\ 
DiReCT~\cite{direct}  & MIMIC-IV   & 5     & ×   & ×     & ×   & en   & 511     \\ \midrule   
ClinicalMC (Ours)       & \begin{tabular}[c]{@{}l@{}}MedEureka,\\ PMC-patients\end{tabular}          & \begin{tabular}[c]{@{}l@{}}16,\\ 24\end{tabular}                 & \checkmark     & \checkmark     & \checkmark      &  \begin{tabular}[c]{@{}l@{}}zh, \\en\end{tabular}          & \begin{tabular}[c]{@{}l@{}}1,275(zh),\\ 5,804(en)\end{tabular} \\ \bottomrule
\end{tabular}
}
\label{compare_data}
\end{table*}

Clinical decision-making is a multi-stage, iterative process that often spans several treatment courses~\cite{MIMICCDM}. 
Upon admission, clinicians first determine the most suitable department for each patient based on their primary presenting symptoms.
During the first course, they gather relevant clinical information and recommend necessary examinations to guide preliminary diagnostic and treatment decisions. 
If the patient’s condition fails to improve, additional examinations are conducted in subsequent courses to reassess the clinical condition and promptly adjust the treatment plan. 
This iterative process continues until the patient’s condition stabilizes and discharge criteria are met.
The overall process is illustrated in Fig.~\ref{workflow}.

Several benchmarks have been proposed for clinical decision-making, which can broadly be categorized into exam-based and clinical case-based benchmarks.
Exam-based benchmarks, such as MedQA~\cite{medqa}, MedMCQA~\cite{Medmcqa}, PubMedQA~\cite{pubmedqa}, and MMLU~\cite{mmlu}, primarily consist of Q\&A pairs extracted from medical books and literature, aiming to evaluate the domain knowledge of LLMs. However, they are largely biased toward theoretical knowledge and fail to align with actual clinical decision scenarios.
Clinical case-based benchmarks such as Clinicallab~\cite{clinicallab}, AI Hospital~\cite{AIHospital}, and MedJourney~\cite{MedJourney} aim to simulate real-world clinical scenarios. 
However, they typically focus on single-course decision-making, involving only a single round of diagnosis and treatment, overlooking the crucial process of reassessing and adjusting treatment plans when a patient fails to improve across multiple courses.
In this work, we further address this gap by modeling multi-course decision-making scenarios that better reflect real clinical practice. 
For ease of comparison, we summarize the differences between our benchmark and the most relevant clinical benchmarks in Table~\ref{compare_data}.

Hence, in the paper, we introduce \textbf{ClinicalMC}, a novel benchmark for evaluating the multi-course clinical decision-making capabilities of LLMs.
To construct this benchmark, we collect clinical records that encompass multiple changes in patient conditions and incorporate condition assessment tasks into each key decision point throughout the clinical course.
In addition, we design a three-round annotation workflow to ensure high-quality and consistent annotations.
Using this approach, we build 1,275 Chinese samples (covering 16 departments) and 5,804 English samples (covering 24 departments) from MedEureka~\cite{medeureka} and PMC-patients~\cite{pmcpatients}.
To facilitate systematic evaluation on ClinicalMC, we develop a multi-agent evaluation framework comprising a patient agent, an examiner agent, and a doctor agent.
The patient agent provides the primary symptoms. 
The examiner agent provides feedback on the examination results. 
The doctor agent makes decisions at each stage of the workflow based on the patient’s evolving condition. 
Using this benchmark and framework, we construct two experimental settings—a single-turn static setting and a multi-turn dynamic setting—and conduct a comprehensive evaluation with a range of doctor agents, including closed-source LLMs such as GPT-4o-mini~\cite{Gpt-4o}, open-source LLMs such as DeepSeek-V3.2~\cite{liu2025DeepSeek}, and medical LLMs such as HuatuoGPT-o1~\cite{huatuogpt2}.

In summary, our contributions include:
\begin{itemize}
    \item We introduce a novel benchmark for multi-course clinical decision-making, ClinicalMC. The benchmark comprises 1,275 Chinese samples across 16 departments and 5,804 English samples across 24 departments.
    \item 
    The main characteristic of ClinicalMC is its inclusion of multiple clinical courses for each patient, enabling a more realistic representation of how a patient’s condition evolves over time. 
    In the English dataset, patients have an average of 5.11 clinical courses, whereas in the Chinese dataset, the average is 3.42.
    
    
    \item 

    We evaluate medical LLMs as well as closed- and open-source LLMs on ClinicalMC, indicating that state-of-the-art medical models like instruction-tuned HuatuoGPT-o1(7B) achieve average performance of 43.40\% and 47.77\% on Chinese and English, respectively, far below human performance (85.00\% and 87.51\%). We further provide detailed analyses and suggest future research directions.
    
    
\end{itemize}

\section{Problem Formulation}

In this work, we evaluate the complete clinical process from patient admission to discharge. Each clinical task can be formally defined as:



\textbf{Triage ($TR$):} 
This task requires the doctor to select the most suitable department $dp$ from a set of candidate departments $ds$, given the patient’s chief complaint $cc$ and basic information $bi$. Formally, this is represented as:
 $dp = TR(cc, bi, ds)$.


\textbf{Examination Recommendation ($ER$):}
This task involves predicting the necessary auxiliary examinations $ex$ based on the patient’s chief complaint, present history $ph_1$, past history $ph_2$, and physical examination $pe$.
Formally, this can be represented as: $ex = ER(cc, bi, ph_1, ph_2, pe, dp)$. 
For examination recommendations across multiple courses, the input includes the patient’s chief complaint of the current course, along with all prior patient information. 
This can be represented as: $ex' = ER(emr, pc, cc', pe')$, where $cc'$, $ex'$, and $pe'$ represent the chief complaint, examination recommendation, and physical examination in the current course. 
$pc$ and $emr$ represent the previous course and the patient's admission information.


\textbf{Clinical Diagnosis ($CD$):}
This task requires the doctor to determine the patient’s preliminary diagnosis $pd$, the corresponding diagnostic basis $pb$, and the differential diagnosis $dd$, based on the patient’s chief complaint, present history, past history, physical examination, and auxiliary examinations. 
It can be formally represented as: $pd, pb, dd = CD(cc, bi, ph_1, ph_2, pe, dp, ex)$.

\textbf{Assessment ($AS$):}
This task requires the doctor to assess the patient’s condition, based on the chief complaint and physical examination of the current course. 
The assessment may involve updating an existing diagnosis or introducing a new one.
Formally, the task is defined as: $as’ = AS(cc', pe', ex', emr)$,
where $as'$ represent the clinical assessment for the current course.


\textbf{Treatment Planning ($TP$):}
This task involves predicting the optimal treatment plan based on the patient’s chief complaint, present history, past history, physical examination, auxiliary examinations, preliminary diagnosis, diagnostic basis, and differential diagnosis. 
It can be formally expressed as: $ tp = TP(emr)$.
For treatment planning across multiple courses, the input also includes the current course’s data.
This can be represented as $tp' = TP(emr, cc', pe', ex', as')$.

\textbf{Final Diagnosis ($FD$):}
This task requires the doctor to determine the final diagnosis $fd$ and its supporting basis $fb$ based on the entire clinical trajectory.
This task can be formally represented as: $fd,fb = FD(emr, pn)$, where $pn = [pc_1, pc_2, \ldots, pc_n]$ is the sequence of $n$ courses.
Each course $pc_i(1\leq i\leq n)$ includes the chief complaint, physical examination, auxiliary examination, assessment, and treatment plan: $pc_i = (cc', ex', pe', as', tp')$.

\section{ClinicalMC Construction}

\label{benchmark}

In this section, we provide a detailed description of the data collection and processing, quality control, and data statistics and analysis.

\begin{figure*}[ht]
\centering
\includegraphics[width=\textwidth]{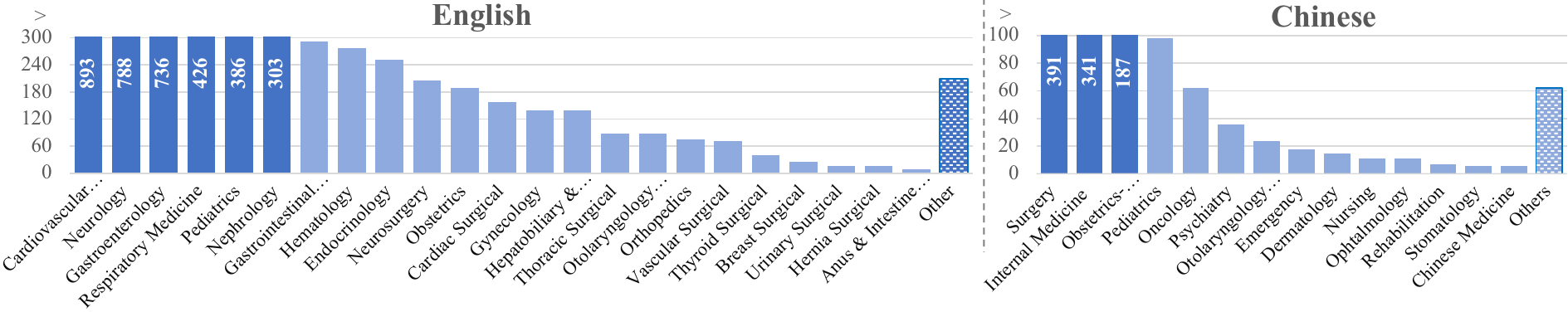} 
\caption{The department distribution of the Chinese and English datasets.}
\label{fenbu}
\end{figure*}

\subsection{Data Collection and Processing}
\label{collect}



For the \textbf{Chinese data}, we use Electronic Health Records (EHRs) from MedEureka as the original data source. 
To obtain strictly anonymized and high-quality EHRs, we process the data in two stages.
In the first stage, we identify EHRs containing personal information (e.g., names, phone numbers) using regular expressions, and replace sensitive data with placeholders (e.g., ``Patient A'') or randomly generated values, resulting in 6,947 EHRs.
In the second stage, we further filter the data to retain only complete and high-quality EHRs. 
We first remove EHRs lacking key information (e.g., chief complaints, diagnoses, or treatment processes), retaining 5,106 EHRs. We then exclude EHRs with a final outcome of death, leaving 4,179 EHRs, and finally eliminate duplicate records through fine-grained demographic matching (e.g., gender and occupation). 
After this stage, we obtain 3,317 high-quality EHRs, each containing multiple treatment courses.
For the \textbf{English data}, we use 167,034 anonymized case reports from PMC-Patients as the original data source. 
To obtain high-quality multi-course reports, we conduct three screening steps.
First, we use the GPT-4o model to remove reports that lack multiple courses or contain incomplete clinical courses (e.g., no improvement or death), retaining 37,357 reports.
Second, we remove reports missing key fields such as admission and final diagnosis, or those labeled as ``undiagnosed'', leaving 15,572 reports.
Finally, we exclude non-human data (e.g., treatment reports for animals).
After this rigorous screening process, we ultimately retain 6,748 reports.
Additionally, to ensure compliance with ethical standards, three clinicians from a Grade 3A hospital conduct a thorough ethical review of the final dataset, confirming that no ethical or moral guidelines are violated.

\begin{table}[t]
\small
\centering
\caption{Statistics of our constructed dataset.}
\resizebox{\columnwidth}{!}{
\begin{tabular}{lccccc}
\toprule
\textbf{Language}  &  \textbf{Avg. Notes} & \textbf{ Max Notes} & \textbf{ Min Notes} & \textbf{\# EHRs} \\ \midrule
English    &  5.11         & 11         & 2            & 5,804 \\
Chinese    &  3.42         & 10         & 2            & 1,275 \\ \bottomrule
\end{tabular}
}
\label{statics}
\end{table}

\subsection{Quality Control}
\label{data_quality}
To construct ClinicalMC, we assemble a professional annotation team comprising three inspectors and two reviewers. 
The dataset is first automatically segmented from multi-course EHRs using an LLM.
Subsequently, three clinically trained inspectors perform an initial verification, followed by a dual review conducted by two senior clinicians. 
The detailed annotation workflow is provided in Appendix~\ref{annotation}. 
After a rigorous two-stage quality review, we obtain 1,275 high-quality Chinese EHRs and 5,804 high-quality English EHRs.
To further ensure data integrity and clinical relevance, we conduct an additional quality-control procedure involving three senior clinicians, each with over ten years of clinical experience and independent of the annotation reviewers.
For this assessment, we randomly sample 3,000 cases from the English dataset (51.68\%) and 1,000 cases from the Chinese dataset (78.43\%). 
We design a standardized scoring framework that presents complete case information and requires clinicians to assess six binary quality dimensions: 1) rationality of course segmentation, 2) accuracy of triage, 3) correctness of diagnostic results, 4) appropriateness of treatment plans, 5) accuracy of clinical assessments, and 6) accuracy of examination recommendations. 
Clinicians make a ``yes/no'' judgment for each dimension, and a case is deemed valid only when all criteria are satisfied.
Evaluation results show that 93.3\% of sampled cases meet the predefined quality standards. 
The pass rates for individual criteria range from 91.9\% to 96.3\%, indicating consistently high overall quality.
The Cohen’s kappa~\cite{banerjee1999beyond} for inter-reviewer agreement is 0.85, demonstrating strong consistency among reviewers. 
For the remaining 6.7\% of cases that do not meet the standards, we perform manual corrections to ensure the reliability and completeness of the dataset.

\subsection{Data Statistics and Analysis}
We conduct an in-depth statistical analysis of clinical decision-making from two perspectives.
\textbf{\textit{1) Department distribution}}.
Fig.~\ref{fenbu} presents the department distribution in both the Chinese and English EHR datasets. 
In the English dataset, the ``Cardiovascular Medicine'' department contains the most samples (893 EHRs), whereas the ``Anus \& Intestine Surgery'' department has the fewest (10 EHRs).
In the Chinese dataset, the ``Surgery'' department has the largest sample size (391 EHRs), while the ``Chinese Medicine'' and ``Stomatology'' department have the smallest, with only 6 EHRs each.
By analyzing the department distribution in both datasets, we observe an imbalance in the sample sizes, reflecting the real-world situation in clinical data. 
This imbalance is likely due to differences in clinical demand across different departments.
\textbf{\textit{2) Number of courses}}.
As shown in Table~\ref{statics}, the English dataset has an average of 5.11 courses per patient, ranging from 2 to 11. In comparison, the Chinese dataset has an average of 3.42 courses per patient, with a range of 
2 to 10.

\begin{figure*}[t]
\centering
\includegraphics[width=\textwidth]{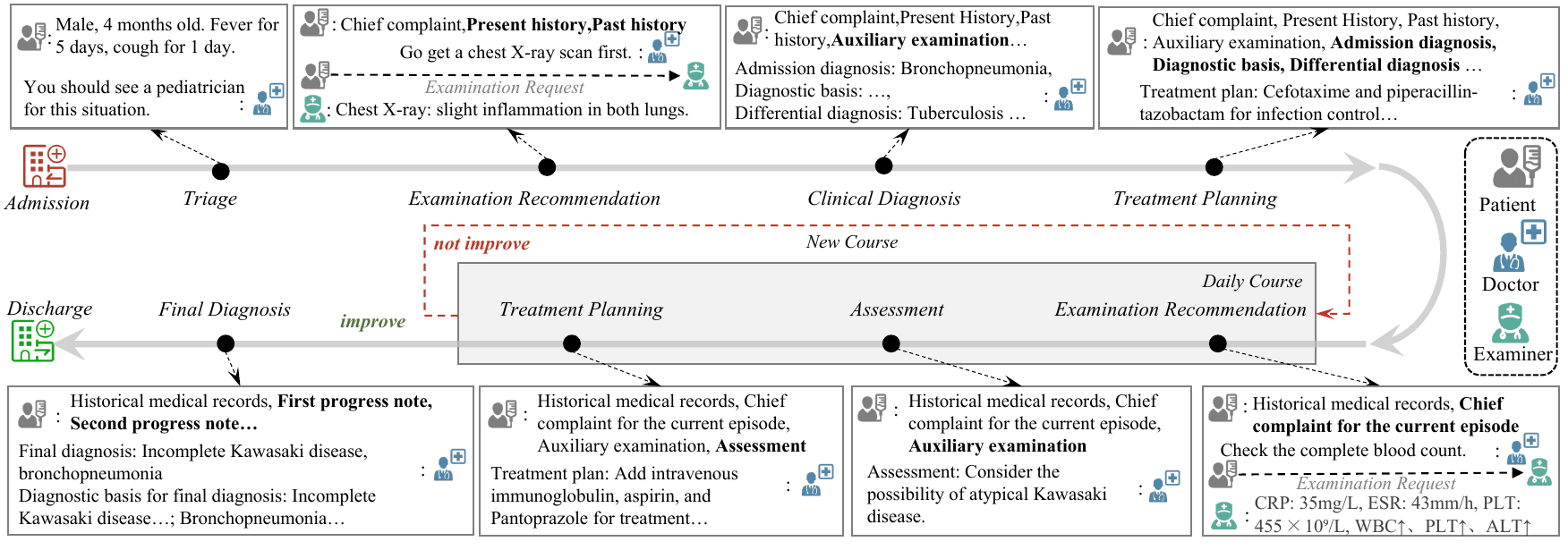} 
\caption{The SimHospital framework includes a doctor agent, an examiner agent, and a patient agent. 
In different tasks, different roles will engage in dialogues. 
When the patient shows improvement and is ready for discharge, the final diagnosis task is performed; otherwise, the patient continues into a new course. 
The \textbf{bold} text indicates the information that has been newly added in the current task compared to the previous task.
}
\label{framework}
\end{figure*}

\section{Evaluation Framework}


Inspired by AI Hospital~\cite{AIHospital}, we develop an evaluation framework, \textbf{SimHospital}, which consists of a doctor agent, a patient agent, and an examiner agent. 
GPT-4o-mini is used for the patient and examiner agents, while various LLMs are employed as the doctor agent to assess clinical decision-making performance.
We also conduct an ablation study of the patient and examiner models in Appendix~\ref{analysis_examiner}.



\subsection{Agent Behavior Setting for All Role}
\textbf{Examiner.} The examiner agent is responsible for providing relevant examination results upon request from the patient agent.
If the requested examination has corresponding results available, the examiner agent returns those results to the doctor. Otherwise, it responds with an indication that no such examination has been conducted.
\textbf{Patient.} The patient agent’s main task is to interact with the doctor and the examiner agents. To match the actual situation, we add the chief complaint, present history, past history, and physical examination to the prompts of the patient agent, but do not specify the diagnosis or treatment plan. If the doctor suggests performing a specific examination, the patient agent follows the suggestion and provides the examination to the examiner agent. 
\textbf{Doctor.} The doctor agent’s primary task is to gather and analyze patient information to complete clinical decision-making tasks, including triage, examination recommendation, clinical diagnosis, assessment, treatment planning, and final diagnosis.




\subsection{Clinical Workflow}
The SimHospital Framework simulates the entire process from admission to recovery and discharge by constructing multiple agents, as illustrated in Fig.~\ref{framework}. 
The interaction begins with the patient agent and proceeds through four stages.
In the first stage, the patient agent presents a chief complaint, and the doctor agent recommends the appropriate department.
In the second stage, the doctor interacts with both the patient and examiner agents, recommends necessary examinations, and makes clinical diagnoses and treatment plans.
In the third stage, the patient agent enters the multi-course phase, during which the doctor agent sequentially performs tasks such as examination recommendation, assessment, and treatment planning based on the patient's complaints for the current course.
This process repeats until the patient recovers and is ready for discharge. 
In the fourth stage, the doctor agent provides a discharge diagnosis based on the patient’s 
complete medical information.

\begin{table*}[ht]
\centering
\caption{Experimental results on English data (\%). ``T'', ``E'', ``PD'', ``PB'', ``DD'', ``TP'', ``FD'', and ``FB'' refer to triage, examination recall, preliminary diagnosis, preliminary diagnosis basis, differential diagnosis, treatment planning, final diagnosis, and final diagnosis basis, respectively. ``CE'', ``CA'', and ``CT'' represent the examination recommendation, assessment, and treatment planning for each course, respectively. 
}
\resizebox{\textwidth}{!}{
\begin{tabular}{
p{3.8cm}
>{\centering}p{1.3cm}
>{\centering}p{1.5cm}
>{\centering}p{1.2cm}
>{\centering}p{1.7cm}
>{\centering}p{1.7cm}
>{\centering}p{1.7cm}
>{\centering}p{1.7cm}
>{\centering}p{1.65cm}
>{\centering}p{1.7cm}
>{\centering}p{1.0cm}
>{\centering}p{1.6cm}
<{\centering}p{0.9cm}
<{\centering}}
\toprule
\textbf{Model} & \textbf{T\_Acc} & \textbf{E\_Recall} & \textbf{PD\_F1} & \textbf{PB\_Score} & \textbf{DD\_Score} & \textbf{TP\_IoU} & \textbf{CE\_Recall} & \textbf{CA\_IoU} & \textbf{CT\_IoU} & \textbf{FD\_F1} & \textbf{FB\_Score} & \textbf{Avg}   \\ \midrule

\multicolumn{13}{c}{\cellcolor[HTML]{EFEFEF}\textit{\textbf{Medical LLMs}}} \\ \midrule
Apollo2-7B & 61.06 & 72.76 & 29.93 & 61.99 & 40.95 & 5.37 & 37.13 & 23.98 & 10.06 & 65.27 & 74.61 & 43.92 \\
Asclepius-Llama2-13B & 0.02 & 0.00 & 0.00 & 44.05 & 38.30 & 1.42 & 0.00 & 22.66 & 1.02 & 0.00 & 31.25 & 12.61 \\
Asclepius-Llama2-7B & 0.02 & 0.00 & 0.00 & 43.65 & 38.45 & 1.42 & 0.00 & 22.49 & 1.01 & 0.00 & 31.29 & 12.58 \\
MedGemma & 63.02 & 19.88 & 22.40 & 76.99 & 62.79 & 10.07 & 24.66 & 50.80 & 2.10 & 70.38 & 85.85 & 44.45 \\
Baichuan-M2 & 61.73 & 25.50 & 24.82 & 74.85 & 51.76 & 9.36 & 25.47 & 54.02 & 2.27 & 80.16 & 86.33 & 45.12 \\
HuatuoGPT-o1-7B & 58.55 & 59.00 & 20.01 & 58.44 & 51.30 & 8.21 & 58.41 & 51.67 & 1.64 & 76.13 & 82.10 & 47.77 \\

\midrule
\multicolumn{13}{c}{\cellcolor[HTML]{EFEFEF}\textit{\textbf{Open-source LLMs}}} \\ \midrule
Llama-3.3-70B & 63.66 & 17.33 & 20.51 & 71.97 & 51.83 & 10.87 & 15.60 & 70.16 & 5.74 & 82.19 & 79.81 & 44.52 \\
Llama-3.2-3B & 46.36 & 34.77 & 14.85 & 48.76 & 38.91 & 5.96 & 26.15 & 48.05 & 3.74 & 63.76 & 69.87 & 36.47 \\
Mistral-7B-v0.3 & 37.35 & 59.37 & 16.47 & 65.53 & 46.47 & 8.45 & 45.89 & 58.97 & 4.85 & 72.96 & 79.37 & 45.06 \\
Mixtral-8x22B & 59.84 & 42.64 & 27.67 & 71.29 & 50.60 & 10.91 & 39.70 & 70.53 & 6.25 & 87.84 & 80.49 & 49.80 \\
Falcon3-7B & 50.76 & 43.81 & 14.67 & 60.57 & 47.12 & 7.47 & 42.08 & 65.48 & 4.36 & 67.57 & 76.51 & 43.67 \\
Qwen2.5-72B & 62.67 & 17.73 & 27.27 & 74.57 & 50.66 & 11.10 & 22.92 & 59.08 & 5.57 & 77.49 & 87.56 & 45.15 \\
Qwen2.5-32B & 62.79 & 19.06 & 22.13 & 74.42 & 52.38 & 11.72 & 24.72 & 66.11 & 6.39 & 79.75 & 86.26 & 45.98 \\
Qwen2.5-14B & 65.13 & 14.98 & 25.85 & 73.88 & 52.28 & 10.40 & 19.18 & 61.67 & 4.63 & 90.59 & 80.61 & 45.38 \\
Qwen2.5-7B & 59.30 & 52.15 & 10.94 & 59.79 & 49.03 & 8.13 & 39.32 & 55.46 & 5.36 & 77.36 & 77.98 & 44.98 \\

Qwen3-Next-80B-A3B & 57.15 & 22.96 & 29.67 & 77.91 & 57.23 & 12.40 & 17.16 & 59.05 & 4.12 & 79.56 & 84.35 & 45.59 \\
DeepSeek-V3.2-Chat & 62.90 & 20.68 & 27.69 & 81.82 & 60.60 & 10.95 & 23.49 & 42.87 & 5.28 & 69.62 & 86.28 & 44.74 \\
DeepSeek-V3.2-Reason & 53.11 & 13.95 & 28.08 & 78.43 & 56.17 & 12.89 & 13.04 & 57.68 & 5.88 & 70.85 & 87.17 & 43.39 \\


\midrule
\multicolumn{13}{c}{\cellcolor[HTML]{EFEFEF}\textit{\textbf{Closed-source LLMs}}} \\ \midrule
GPT-4o-mini & 65.73 & 17.74 & 34.73 & 72.80 & 59.08 & 11.62 & 17.16 & 54.44 & 2.30 & 93.13 & 79.75 & 46.23 \\
GPT-5-mini & 61.71 & 28.16 & 31.86 & 83.33 & 61.71 & 13.20 & 25.84 & 60.01 & 4.71 & 58.46 & 86.65 & 46.88 \\
Qwen-turbo & 63.78 & 42.24 & 32.96 & 72.30 & 55.08 & 9.97 & 51.41 & 46.68 & 2.09 & 89.96 & 79.98 & 49.68 \\
\midrule

\multicolumn{13}{c}{\cellcolor[HTML]{EFEFEF}\textit{\textbf{Other Method}}}           \\ \midrule
Human (sampling)       & 90.00     & 86.39    & 88.75    & 84.85    & 82.80    & 83.06      & 83.05   & 87.22      & 92.09      & 90.95     & 93.45      & \textbf{87.51} \\ 
\bottomrule
\end{tabular}}
\label{main_results_en}
\end{table*}

\section{Experiments}
In this section, we implement state-of-the-art models on our newly constructed ClinicalMC benchmark, aiming at assessing their performance and identifying the underlying challenges.

\subsection{Experimental Setup}
\label{setup}
\textbf{Baseline Model.} 
We evaluate four categories of LLMs:
1) \textbf{Medical LLMs,} including MedGemma~\cite{Medgemma}, Baichuan-M2~\cite{baichuanm2}, HuatuoGPT-o1, and Apollo2-7B~\cite{apollo2}.
Additionally, as some medical LLMs demonstrate strong performance across different languages, we use HuatuoGPT2 (7B, 13B, and 34B)~\cite{huatuogpt2} for Chinese datasets and Asclepius-Llama2 (7B and 13B)~\cite{Asclepius} for English datasets. 
Notably, Asclepius-Llama2 was trained on the PMC-Patients dataset, making it well-suited for \textbf{assessing potential data leakage risks}.
2) \textbf{Open-source LLMs}, including Falcon3-7B~\cite{falcon}, Qwen2.5 (ranging from 7B to 72B), DeepSeek-V3.2 (Chat and Reason)~\cite{liu2025DeepSeek}, Llama-3.3-70B~\cite{llama3}, Llama-3.2-3B, Mistral-7B~\cite{mistral7b}, Mixtral-8x22B~\cite{mixtraloe}, and Qwen3-Next-80B-A3B\footnote{\url{https://huggingface.co/Qwen/Qwen3-Next-80B-A3B-Instruct}}.
Since the Llama series models exhibit certain capabilities in processing Chinese, we also evaluate their performance on Chinese datasets. 
3) \textbf{Closed-source LLMs}, such as GPT-4o-mini~\cite{Gpt-4o}, GPT-5-mini, and Qwen-turbo. 
4) \textbf{Other Method}. 
We randomly select 100 samples and invite a medical student who does not participate in the data annotation process to answer the questions.

\begin{table*}[ht]
\centering
\caption{Experimental results on Chinese data (\%).}
\resizebox{\textwidth}{!}{
\begin{tabular}{
p{3.8cm}
>{\centering}p{1.3cm}
>{\centering}p{1.5cm}
>{\centering}p{1.2cm}
>{\centering}p{1.7cm}
>{\centering}p{1.7cm}
>{\centering}p{1.7cm}
>{\centering}p{1.7cm}
>{\centering}p{1.65cm}
>{\centering}p{1.7cm}
>{\centering}p{1.0cm}
>{\centering}p{1.6cm}
<{\centering}p{0.9cm}
<{\centering}}
\toprule
\textbf{Model} & \textbf{T\_Acc} & \textbf{E\_Recall} & \textbf{PD\_F1} & \textbf{PB\_Score} & \textbf{DD\_Score} & \textbf{TP\_IoU} & \textbf{CE\_Recall} & \textbf{CA\_IoU} & \textbf{CT\_IoU} & \textbf{FD\_F1} & \textbf{FB\_Score} & \textbf{Avg} \\ \midrule

\multicolumn{13}{c}{\cellcolor[HTML]{EFEFEF}\textit{\textbf{Medical LLMs}}} \\ \midrule
Apollo2-7B & 52.71 & 29.05 & 33.85 & 66.65 & 35.01 & 5.40 & 29.39 & 46.54 & 0.98 & 25.82 & 54.75 & 34.56 \\
HuatuoGPT2-7B & 42.82 & 27.72 & 0.26 & 49.80 & 29.99 & 2.47 & 40.79 & 28.83 & 0.46 & 0.47 & 24.36 & 22.54 \\
HuatuoGPT2-13B & 45.13 & 26.63 & 0.00 & 48.26 & 29.87 & 1.99 & 52.94 & 23.99 & 0.06 & 0.00 & 37.69 & 24.23 \\
HuatuoGPT2-34B & 57.59 & 29.65 & 26.18 & 61.53 & 34.43 & 5.04 & 42.84 & 35.31 & 0.73 & 22.24 & 37.23 & 32.07 \\
MedGemma & 65.65 & 28.64 & 30.13 & 76.78 & 46.34 & 6.91 & 13.73 & 79.74 & 2.92 & 58.48 & 83.72 & 44.82 \\
HuatuoGPT-o1-7B & 65.73 & 30.93 & 33.31 & 70.48 & 39.36 & 5.78 & 17.24 & 71.77 & 2.89 & 62.55 & 77.38 & 43.40 \\
Baichuan-M2 & 54.20 & 32.74 & 35.67 & 78.37 & 47.03 & 5.11 & 18.81 & 76.34 & 2.80 & 68.45 & 80.93 & 45.50 \\

\midrule
\multicolumn{13}{c}{\cellcolor[HTML]{EFEFEF}\textit{\textbf{Open-source LLMs}}} \\ \midrule
Llama-3.3-70B & 52.39 & 35.52 & 33.72 & 73.04 & 41.11 & 6.38 & 25.28 & 50.81 & 1.97 & 29.12 & 67.94 & 37.93 \\
Llama-3.2-3B & 46.20 & 23.79 & 10.63 & 38.32 & 24.33 & 1.84 & 50.67 & 21.55 & 0.18 & 3.47 & 21.13 & 22.01 \\
Mistral-7B-v0.3 & 39.14 & 23.90 & 14.52 & 34.15 & 22.78 & 2.28 & 24.60 & 23.67 & 0.49 & 13.84 & 27.75 & 20.65 \\
Mixtral-8x22B & 53.10 & 27.23 & 26.15 & 60.19 & 25.16 & 4.75 & 22.97 & 36.26 & 1.14 & 23.75 & 52.00 & 30.25 \\
Falcon3-7B & 41.10 & 29.05 & 12.60 & 34.78 & 22.71 & 1.47 & 37.09 & 23.62 & 0.41 & 6.93 & 24.85 & 21.33 \\
Qwen2.5-72B-Chat & 57.41 & 23.74 & 37.96 & 73.98 & 40.05 & 5.94 & 20.88 & 49.17 & 2.11 & 31.88 & 70.32 & 37.59 \\
Qwen2.5-32B-Chat & 59.92 & 23.06 & 34.98 & 76.61 & 43.45 & 7.67 & 18.24 & 60.09 & 2.84 & 30.69 & 64.42 & 38.36 \\
Qwen2.5-14B-Chat & 53.57 & 21.16 & 36.20 & 74.73 & 37.99 & 7.84 & 23.68 & 63.22 & 2.63 & 30.54 & 64.36 & 37.81 \\
Qwen2.5-7B-Chat & 62.35 & 28.87 & 35.35 & 72.49 & 38.53 & 6.34 & 31.33 & 50.66 & 1.40 & 28.90 & 64.75 & 38.27 \\
Qwen3-Next-80B-A3B & 68.31 & 28.53 & 31.25 & 82.46 & 61.30 & 6.38 & 11.98 & 84.89 & 2.75 & 63.80 & 88.17 & 48.17 \\
DeepSeek-V3.2-Chat & 67.06 & 25.46 & 30.87 & 77.04 & 63.44 & 7.57 & 15.55 & 77.94 & 3.07 & 54.61 & 79.22 & 45.62 \\
DeepSeek-V3.2-Reason & 59.50 & 23.94 & 37.98 & 77.10 & 58.65 & 6.86 & 13.67 & 81.02 & 3.24 & 67.36 & 81.08 & 46.40 \\

\midrule
\multicolumn{13}{c}{\cellcolor[HTML]{EFEFEF}\textit{\textbf{Closed-source LLMs}}} \\ \midrule
GPT-4o-mini & 54.43 & 20.65 & 32.58 & 69.47 & 36.41 & 5.81 & 15.46 & 53.97 & 1.56 & 28.28 & 64.66 & 34.84 \\
GPT-5-mini & 59.62 & 25.92 & 12.00 & 83.76 & 49.48 & 6.76 & 9.66 & 84.34 & 4.19 & 44.60 & 82.16 & 42.04 \\
Qwen-turbo & 55.69 & 25.88 & 33.80 & 73.29 & 38.68 & 7.49 & 26.13 & 62.31 & 2.71 & 29.77 & 69.08 & 38.62 \\ \midrule

\multicolumn{13}{c}{\cellcolor[HTML]{EFEFEF}\textit{\textbf{Other Method}}}           \\ \midrule
Human (sampling)       & 90.91     & 85.61    & 87.68    & 82.91    & 88.14    & 79.41      & 86.70   & 82.76      & 84.53      & 83.71     & 82.73      & \textbf{85.00} \\ 
\bottomrule

\end{tabular}}
\label{main_results_zh}
\end{table*}


\textbf{Evaluation Metrics.} 
For the triage task, we use $Accuracy (Acc)$ as the evaluation metric. 
For examination recommendation, we adopt $Recall$, and for disease diagnosis—covering both clinical and final diagnosis—we use the $F1$ score.
To evaluate examination and diagnosis entities, we construct a standardized synonym list by first collecting synonyms from the Medeureka\_corpus~\cite{medeureka}.
To address terminological inconsistencies due to differences in model training data, we select the largest model from each of 10 LLM series to independently generate synonym lists. 
These are merged with the Medeureka corpus and refined by three clinicians to ensure consistency and clinical validity.
For diagnosis basis, we employ an LLM to assess (1) whether the medical reasoning process is logically coherent, and (2) whether the provided evidence sufficiently and effectively supports the predicted diagnosis. 
The scores for preliminary and final diagnosis are denoted as $PB\_Score$ and $FB\_Score$, respectively. The detailed evaluation prompts are provided in Appendix~\ref{example}.
For differential diagnosis, we focus on evaluating whether the set of predicted differential diagnoses adequately covers other clinically significant potential conditions. This metric is denoted as $DD\_Score$. The detailed evaluation prompts are provided in Appendix~\ref{example}.
For the assessment and treatment planning in each clinical course, we follow the approach of MedChain~\cite{medchain} by decomposing model outputs into structured clinical entities. 
We then compute the Intersection over Union (IoU) between these entities and the gold-standard key interventions. This metric emphasizes the coverage of critical clinical actions rather than surface-level wording, and is denoted as $CA\_IoU$ and $CT\_IoU$, respectively.



\textbf{Implementation Details.} 
We design two experimental settings: a single-round static setting and a multi-round dynamic setting. 
In the former, the ground-truth annotations from preceding tasks are provided as inputs to subsequent tasks. 
In the latter, the model outputs of preceding tasks are used as inputs to subsequent tasks. Detailed implementation details are presented in Appendix~\ref{implementation}.



\subsection{Main Results}
\label{single_turn}

We systematically evaluate all baseline LLMs on the ClinicalMC under the single-turn setting. 
The English and Chinese results are reported in Table~\ref{main_results_en} and Table~\ref{main_results_zh}, respectively.
We select one representative model from each of the medical, closed-source, and open-source LLM categories for evaluation under the multi-turn experimental setting. The detailed results are presented in Appendix~\ref{multi-turn}.



From the Table~\ref{main_results_en} and Table~\ref{main_results_zh}, we notice that: 
1) All LLMs perform poorly on both the English and Chinese datasets, leaving substantial room for improvement compared to human performance ($Avg$ scores of 85.00\% and  87.51\%, respectively). 
The best-performing model achieves only 49.68\% on the English dataset and 48.17\% on the Chinese dataset, highlighting the significant challenge posed by our ClinicalMC benchmark.
2) LLMs perform worse in multi-course settings compared to single-course settings. 
Specifically, Llama-3.3-70B achieves a $TP\_IoU$ of 6.38\% on Chinese data and 10.87\% on English data, outperforming the $CT\_IoU$ by 4.41\% and 5.13\%, respectively.
This decline is primarily due to the increasing complexity of clinical information as the number of courses grows. 
Patients’ records often contain redundant or repeated examinations and treatments, making it more challenging for the model to accurately assess the current condition and generate up-to-date treatment plans in real-time.
3) Notably, although Asclepius-Llama2 is trained on the PMC-Patients dataset, it performs poorly on the English subset of ClinicalMC. Specifically, the 7B and 13B variants  achieve $Avg$ of only 12.58\% and 12.61\%, respectively. 
This is primarily because ClinicalMC reconstructs the medical records into reasoning tasks that require cross-trajectory information integration and explicit clinical decision-making, through multi-trajectory decomposition and multiple rounds of human review. 
In contrast, Asclepius-Llama2 focuses more on medical record generation and local semantic modeling, limiting its effectiveness in such complex clinical reasoning scenarios.
Consequently, these results further highlight the challenging nature of ClinicalMC for evaluating clinical reasoning capabilities.

\subsection{Error Type}
To guide future research in clinical decision-making for LLMs, we manually analyze and classify 200 error samples generated by LLMs on the Chinese and English datasets of ClinicalMC.
These errors are categorized into five types:
(a) \textbf{Redundant Diagnostic and Treatment Plan} (RDTP): The model generates an excessive number of unnecessary diagnostic tests and treatment plans.
(b) \textbf{Failure to Detect Subtle but Critical Changes} (FDSC): 
The model fails to recognize subtle yet clinically significant changes in a patient’s condition—such as slight fluctuations in laboratory results—which may lead to delayed or inappropriate adjustments in diagnosis or treatment plans.
(c) \textbf{Incorrect Clinical Diagnosis} (ICD): 
Due to a lack of domain-specific medical knowledge or misinterpretation of clinical information, the model produces incorrect diagnostic conclusions.
(d) \textbf{Incorrect Reasoning Chain} (IRC): The diagnostic rationale produced by the model does not align with the actual clinical condition. 
(e) \textbf{Other Errors}: all other cases of errors. 
The error distribution is shown in Fig.~\ref{error_distribute}, with illustrative Chinese and English examples included in Appendix~\ref{error_case}.



\begin{figure}[t] 
\centering
\includegraphics[width=0.95\columnwidth]{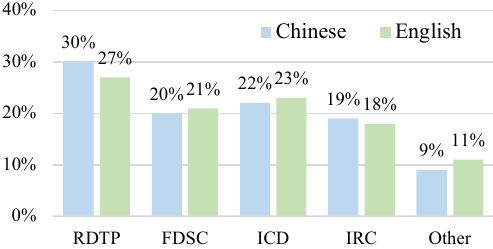} 
\caption{Distribution of error types.}
\label{error_distribute}
\end{figure}

\section{Related Work}

\textbf{Clinical decision-making benchmark}. Clinical decision-making tasks refer to assisting doctors in making the most appropriate diagnosis and treatment decisions by continuously analyzing the patient's chief complaints, medical history, examination results and other information~\cite{MIMICCDM}.
Existing clinical decision-making benchmarks can be broadly classified into two types: exam-based and clinical case-based benchmarks. 
Exam-based benchmarks include datasets such as MedQA~\cite{medqa}, MedMCQA~\cite{Medmcqa}, PubMedQA~\cite{pubmedqa}, and MMLU~\cite{mmlu}, primarily consist of Q\&A pairs extracted from medical books
and literature.
However, there is still a certain gap between these benchmarks and actual clinical decision-making.
Therefore, recent studies have proposed benchmarks based on clinical case benchmarks, such as MedChain~\cite{medchain}, Clinicallab~\cite{clinicallab}, MSDiagnosis~\cite{msdiagnosis}, Ai-Hospital~\cite{AIHospital}, and MedJourney~\cite{MedJourney}. However, these benchmarks mostly focus on a single course or simulate decision-making in outpatient settings. 
For example, Clinicallab evaluates tasks such as department guidance, clinical diagnosis, and treatment planning, but does not involve continuous assessment of patients after treatment until they recover and are discharged. Therefore, this study focuses on the clinical decision-making performance of models in multi-course after patient admission.

\textbf{Agent for medical decision-making}. Research on intelligent agents for medical decision-making can be divided into single-agent~\cite{mmedagent, cod, hou2026cdaflow} and multi-agent~\cite{medagents, agenthospital} methods. 
In single-agent research, CoD~\cite{cod} assesses potential candidate diseases by planning to inquire about the patient's latent symptoms and generates a diagnostic chain from symptoms to possible diseases.
In multi-agent research, medical decision-making problems are typically tackled through a multi-agent task division and collaboration paradigm, such as in frameworks like MDAgents~\cite{mdagents}, MedAgents~\cite{medagents}, and Agent Hospital~\cite{agenthospital}. 
MDAgents is a multi-agent framework that utilizes adaptive decision-making mechanisms to tackle medical decision-making challenges. It operates through multiple phases, including analyzing problem complexity, dynamically recruiting experts, and employing reasoning and decision-making processes at various stages to solve medical Q\&A.

\section{Conclusion}
\label{conclusion}
We introduce ClinicalMC, a benchmark comprising both Chinese and English datasets that encompass the full patient journey from admission to discharge. 
These stages cover triage, first-course examination/diagnosis/treatment, subsequent multi-course examination/assessment/treatment, and final diagnosis. 
To evaluate model performance in multi-course clinical decision-making, we develop a multi-agent framework involving patient, examiner, and doctor agents.
Based on the dataset and framework, we define two experimental settings—single-turn and multi-turn—and evaluate medical LLMs as well as closed-source and open-source LLMs and conduct extensive experimental analysis. 
The results show that ClinicalMC is a challenging dataset that warrants further research and exploration. 


\section*{Limitations}

This paper has two primary limitations that offer avenues for future research:
First, the lack of multimodal information. The raw data used in this study primarily consist of textual medical records collected during hospitalization and do not cover multimodal data across multiple courses, such as medical images and time-series physiological signals. 
In future work, we plan to investigate the integration of medical imaging with text-based reasoning to support clinical decision-making over heterogeneous, multi-source data.
Second, an imbalanced department distribution. The current dataset is mainly derived from a single data source, leading to imbalanced distributions across clinical departments. Although this imbalance partially reflects real-world clinical practice, it may still affect the model’s generalization performance in underrepresented departments. 
In future work, we will incorporate data from multiple healthcare systems to expand coverage and mitigate department-level imbalance.


\section*{Ethical Consideration}

Our ClinicalMC benchmark is based on PMC-Patients and MedEureka, licensed under the Creative Commons Attribution 4.0 License. Accordingly, we assign the copyright of ClinicalMC to the CC-BY 4.0 license. In addition, we have meticulously reviewed our dataset to ensure it does not contain any harmful content, including gender bias, racial discrimination, or inappropriate material.

\section*{Acknowledgments}

This paper was supported by the Shanghai Science and Technology Innovation Action Plan in Computational Biology (No. 24JS2840200).

\bibliography{custom}

@article{pmcpatients,
  title={Pmc-patients: A large-scale dataset of patient summaries and relations for benchmarking retrieval-based clinical decision support systems},
  author={Zhao, Zhengyun and Jin, Qiao and Chen, Fangyuan and Peng, Tuorui and Yu, Sheng},
  journal={arXiv preprint arXiv:2202.13876},
  year={2022}
}

@article{lin2023survey,
  title={A survey on neural data-to-text generation},
  author={Lin, Yupian and Ruan, Tong and Liu, Jingping and Wang, Haofen},
  journal={IEEE Transactions on Knowledge and Data Engineering},
  volume={36},
  number={4},
  pages={1431--1449},
  year={2023},
  publisher={IEEE}
}

@article{liu2025deepseek,
  title={Deepseek-v3. 2: Pushing the frontier of open large language models},
  author={Liu, Aixin and Mei, Aoxue and Lin, Bangcai and Xue, Bing and Wang, Bingxuan and Xu, Bingzheng and Wu, Bochao and Zhang, Bowei and Lin, Chaofan and Dong, Chen and others},
  journal={arXiv preprint arXiv:2512.02556},
  year={2025}
}

@article{hou2026cdaflow,
  title={CDAFlow: Enhancing LLM Clinical Decision-Making through Agentic Workflow},
  author={Hou, Ruihui and Xue, Dongge and Sun, Hongli and He, Ping and Zhang, Weiyan and Ruan, Tong},
  journal={Expert Systems with Applications},
  pages={131806},
  year={2026},
  publisher={Elsevier}
}

@inproceedings{synthetic,
    title = "Publicly Shareable Clinical Large Language Model Built on Synthetic Clinical Notes",
    author = "Kweon, Sunjun  and
      Kim, Junu  and
      Kim, Jiyoun  and
      Im, Sujeong  and
      Cho, Eunbyeol  and
      Bae, Seongsu  and
      Oh, Jungwoo  and
      Lee, Gyubok  and
      Moon, Jong Hak  and
      You, Seng Chan  and
      Baek, Seungjin  and
      Han, Chang Hoon  and
      Jung, Yoon Bin  and
      Jo, Yohan  and
      Choi, Edward",
    editor = "Ku, Lun-Wei  and
      Martins, Andre  and
      Srikumar, Vivek",
    booktitle = "Findings of the Association for Computational Linguistics: ACL 2024",
    month = aug,
    year = "2024",
    address = "Bangkok, Thailand",
    publisher = "Association for Computational Linguistics",
    url = "https://aclanthology.org/2024.findings-acl.305/",
    doi = "10.18653/v1/2024.findings-acl.305",
    pages = "5148--5168"
}

@article{llama,
  title={Llama: Open and efficient foundation language models},
  author={Touvron, Hugo and Lavril, Thibaut and Izacard, Gautier and Martinet, Xavier and Lachaux, Marie-Anne and Lacroix, Timoth{\'e}e and Rozi{\`e}re, Baptiste and Goyal, Naman and Hambro, Eric and Azhar, Faisal and others},
  journal={arXiv preprint arXiv:2302.13971},
  year={2023}
}

@article{mimic3,
  title={MIMIC-III, a freely accessible critical care database},
  author={Johnson, Alistair EW and Pollard, Tom J and Shen, Lu and Lehman, Li-wei H and Feng, Mengling and Ghassemi, Mohammad and Moody, Benjamin and Szolovits, Peter and Anthony Celi, Leo and Mark, Roger G},
  journal={Scientific data},
  volume={3},
  number={1},
  pages={1--9},
  year={2016},
  publisher={Nature Publishing Group}
}

@article{i2b2,
  title={Evaluating the state-of-the-art in automatic de-identification},
  author={Uzuner, {\"O}zlem and Luo, Yuan and Szolovits, Peter},
  journal={Journal of the American Medical Informatics Association},
  volume={14},
  number={5},
  pages={550--563},
  year={2007},
  publisher={BMJ Group BMA House, Tavistock Square, London, WC1H 9JR}
}

@article{Gpt-4o,
  title={Gpt-4o system card},
  author={Hurst, Aaron and Lerer, Adam and Goucher, Adam P and Perelman, Adam and Ramesh, Aditya and Clark, Aidan and Ostrow, AJ and Welihinda, Akila and Hayes, Alan and Radford, Alec and others},
  journal={arXiv preprint arXiv:2410.21276},
  year={2024}
}

@article{MedRBench,
  title={Quantifying the reasoning abilities of llms on real-world clinical cases},
  author={Qiu, Pengcheng and Wu, Chaoyi and Liu, Shuyu and Zhao, Weike and Zhang, Ya and Wang, Yanfeng and Xie, Weidi},
  journal={arXiv preprint arXiv:2503.04691},
  year={2025}
}

@article{clinicallab,
  title={ClinicalLab: Aligning Agents for Multi-Departmental Clinical Diagnostics in the Real World},
  author={Yan, Weixiang and Liu, Haitian and Wu, Tengxiao and Chen, Qian and Wang, Wen and Chai, Haoyuan and Wang, Jiayi and Zhao, Weishan and Zhang, Yixin and Zhang, Renjun and others},
  journal={arXiv preprint arXiv:2406.13890},
  year={2024}
}

@article{MIMICCDM,
  title={Evaluation and mitigation of the limitations of large language models in clinical decision-making},
  author={Hager, Paul and Jungmann, Friederike and Holland, Robbie and Bhagat, Kunal and Hubrecht, Inga and Knauer, Manuel and Vielhauer, Jakob and Makowski, Marcus and Braren, Rickmer and Kaissis, Georgios and others},
  journal={Nature medicine},
  volume={30},
  number={9},
  pages={2613--2622},
  year={2024},
  publisher={Nature Publishing Group US New York}
}

@inproceedings{medchain,
title={MedChain: Bridging the Gap Between {LLM} Agents and Clinical Practice with Interactive Sequence},
author={Jie Liu and Wenxuan Wang and Zizhan Ma and Guolin Huang and SU Yihang and Kao-Jung Chang and Haoliang Li and Linlin Shen and Michael Lyu and Wenting Chen},
booktitle={The Thirty-ninth Annual Conference on Neural Information Processing Systems Datasets and Benchmarks Track},
year={2025},
url={https://openreview.net/forum?id=YvuufwkFJY}
}

@inproceedings{AIHospital,
    title = "{AI} Hospital: Benchmarking Large Language Models in a Multi-agent Medical Interaction Simulator",
    author = "Fan, Zhihao  and
      Wei, Lai  and
      Tang, Jialong  and
      Chen, Wei  and
      Siyuan, Wang  and
      Wei, Zhongyu  and
      Huang, Fei",
    editor = "Rambow, Owen  and
      Wanner, Leo  and
      Apidianaki, Marianna  and
      Al-Khalifa, Hend  and
      Eugenio, Barbara Di  and
      Schockaert, Steven",
    booktitle = "Proceedings of the 31st International Conference on Computational Linguistics",
    month = jan,
    year = "2025",
    address = "Abu Dhabi, UAE",
    publisher = "Association for Computational Linguistics",
    url = "https://aclanthology.org/2025.coling-main.680/",
    pages = "10183--10213"
}

@article{CRAFTMD,
  title={An evaluation framework for clinical use of large language models in patient interaction tasks},
  author={Johri, Shreya and Jeong, Jaehwan and Tran, Benjamin A and Schlessinger, Daniel I and Wongvibulsin, Shannon and Barnes, Leandra A and Zhou, Hong-Yu and Cai, Zhuo Ran and Van Allen, Eliezer M and Kim, David and others},
  journal={Nature Medicine},
  pages={1--10},
  year={2025},
  publisher={Nature Publishing Group US New York}
}

@article{Map,
  title={Map: Evaluation and multi-agent enhancement of large language models for inpatient pathways},
  author={Chen, Zhen and Peng, Zhihao and Liang, Xusheng and Wang, Cheng and Liang, Peigan and Zeng, Linsheng and Ju, Minjie and Yuan, Yixuan},
  journal={arXiv preprint arXiv:2503.13205},
  year={2025}
}

@article{MedJourney,
  title={MedJourney: Benchmark and Evaluation of Large Language Models over Patient Clinical Journey},
  author={Wu, Xian and Zhao, Yutian and Zhang, Yunyan and Wu, Jiageng and Zhu, Zhihong and Zhang, Yingying and Ouyang, Yi and Zhang, Ziheng and Wang, Huimin and Yang, Jie and others},
  journal={Advances in Neural Information Processing Systems},
  volume={37},
  pages={87621--87646},
  year={2024}
}

@article{falcon,
  title={The falcon series of open language models},
  author={Almazrouei, Ebtesam and Alobeidli, Hamza and Alshamsi, Abdulaziz and Cappelli, Alessandro and Cojocaru, Ruxandra and Debbah, M{\'e}rouane and Goffinet, {\'E}tienne and Hesslow, Daniel and Launay, Julien and Malartic, Quentin and others},
  journal={arXiv preprint arXiv:2311.16867},
  year={2023}
}

@article{llama3,
  title={The llama 3 herd of models},
  author={Grattafiori, Aaron and Dubey, Abhimanyu and Jauhri, Abhinav and Pandey, Abhinav and Kadian, Abhishek and Al-Dahle, Ahmad and Letman, Aiesha and Mathur, Akhil and Schelten, Alan and Vaughan, Alex and others},
  journal={arXiv preprint arXiv:2407.21783},
  year={2024}
}

@article{mixtraloe,
  title={Mixtral of experts},
  author={Jiang, Albert Q and Sablayrolles, Alexandre and Roux, Antoine and Mensch, Arthur and Savary, Blanche and Bamford, Chris and Chaplot, Devendra Singh and Casas, Diego de las and Hanna, Emma Bou and Bressand, Florian and others},
  journal={arXiv preprint arXiv:2401.04088},
  year={2024}
}

@misc{mistral7b,
      title={Mistral 7B}, 
      author={Albert Q. Jiang and Alexandre Sablayrolles and Arthur Mensch and Chris Bamford and Devendra Singh Chaplot and Diego de las Casas and Florian Bressand and Gianna Lengyel and Guillaume Lample and Lucile Saulnier and Lélio Renard Lavaud and Marie-Anne Lachaux and Pierre Stock and Teven Le Scao and Thibaut Lavril and Thomas Wang and Timothée Lacroix and William El Sayed},
      year={2023},
      archiveprefix = {arXiv},
      eprint={2310.06825},
      primaryClass={cs.CL},
      url={https://arxiv.org/abs/2310.06825}, 
}

@article{huatuogpt2,
  title={Huatuogpt-ii, one-stage training for medical adaption of llms},
  author={Chen, Junying and Wang, Xidong and Ji, Ke and Gao, Anningzhe and Jiang, Feng and Chen, Shunian and Zhang, Hongbo and Song, Dingjie and Xie, Wenya and Kong, Chuyi and others},
  journal={arXiv preprint arXiv:2311.09774},
  year={2023}
}

@article{apollo2,
  title={Efficiently Democratizing Medical LLMs for 50 Languages via a Mixture of Language Family Experts},
  author={Zheng, Guorui and Wang, Xidong and Liang, Juhao and Chen, Nuo and Zheng, Yuping and Wang, Benyou},
  journal={arXiv preprint arXiv:2410.10626},
  year={2024}
}

@inproceedings{Asclepius,
    title = "Publicly Shareable Clinical Large Language Model Built on Synthetic Clinical Notes",
    author = "Kweon, Sunjun  and
      Kim, Junu  and
      Kim, Jiyoun  and
      Im, Sujeong  and
      Cho, Eunbyeol  and
      Bae, Seongsu  and
      Oh, Jungwoo  and
      Lee, Gyubok  and
      Moon, Jong Hak  and
      You, Seng Chan  and
      Baek, Seungjin  and
      Han, Chang Hoon  and
      Jung, Yoon Bin  and
      Jo, Yohan  and
      Choi, Edward",
    editor = "Ku, Lun-Wei  and
      Martins, Andre  and
      Srikumar, Vivek",
    booktitle = "Findings of the Association for Computational Linguistics: ACL 2024",
    month = aug,
    year = "2024",
    address = "Bangkok, Thailand",
    publisher = "Association for Computational Linguistics",
    url = "https://aclanthology.org/2024.findings-acl.305/",
    doi = "10.18653/v1/2024.findings-acl.305",
    pages = "5148--5168"
}

@article{medqa,
  title={What disease does this patient have? a large-scale open domain question answering dataset from medical exams},
  author={Jin, Di and Pan, Eileen and Oufattole, Nassim and Weng, Wei-Hung and Fang, Hanyi and Szolovits, Peter},
  journal={Applied Sciences},
  volume={11},
  number={14},
  pages={6421},
  year={2021},
  publisher={MDPI}
}

@inproceedings{Medmcqa,
  title={Medmcqa: A large-scale multi-subject multi-choice dataset for medical domain question answering},
  author={Pal, Ankit and Umapathi, Logesh Kumar and Sankarasubbu, Malaikannan},
  booktitle={Conference on health, inference, and learning},
  pages={248--260},
  year={2022},
  organization={PMLR}
}

@article{pubmedqa,
  title={Pubmedqa: A dataset for biomedical research question answering},
  author={Jin, Qiao and Dhingra, Bhuwan and Liu, Zhengping and Cohen, William W and Lu, Xinghua},
  journal={arXiv preprint arXiv:1909.06146},
  year={2019}
}

@article{mmlu,
  title={Measuring Massive Multitask Language Understanding},
  author={Dan Hendrycks and Collin Burns and Steven Basart and Andy Zou and Mantas Mazeika and Dawn Song and Jacob Steinhardt},
  journal={Proceedings of the International Conference on Learning Representations (ICLR)},
  year={2021}
}

@article{msdiagnosis,
  title={MSDiagnosis: An EMR-based Dataset for Clinical Multi-Step Diagnosis},
  author={Hou, Ruihui and Chen, Shencheng and Fan, Yongqi and Zhu, Lifeng and Sun, Jing and Liu, Jingping and Ruan, Tong},
  journal={arXiv preprint arXiv:2408.10039},
  year={2024}
}

@article{cod,
  title={CoD, Towards an Interpretable Medical Agent using Chain of Diagnosis},
  author={Chen, Junying and Gui, Chi and Gao, Anningzhe and Ji, Ke and Wang, Xidong and Wan, Xiang and Wang, Benyou},
  journal={arXiv preprint arXiv:2407.13301},
  year={2024}
}

@inproceedings{mdagents,
    title={{MDA}gents: An Adaptive Collaboration of {LLM}s for Medical Decision-Making},
    author={Yubin Kim and Chanwoo Park and Hyewon Jeong and Yik Siu Chan and Xuhai Xu and Daniel McDuff and Hyeonhoon Lee and Marzyeh Ghassemi and Cynthia Breazeal and Hae Won Park},
    booktitle={The Thirty-eighth Annual Conference on Neural Information Processing Systems},
    year={2024},
    url={https://openreview.net/forum?id=EKdk4vxKO4}
}

@inproceedings{medagents,
    title = "{M}ed{A}gents: Large Language Models as Collaborators for Zero-shot Medical Reasoning",
    author = "Tang, Xiangru  and
      Zou, Anni  and
      Zhang, Zhuosheng  and
      Li, Ziming  and
      Zhao, Yilun  and
      Zhang, Xingyao  and
      Cohan, Arman  and
      Gerstein, Mark",
    editor = "Ku, Lun-Wei  and
      Martins, Andre  and
      Srikumar, Vivek",
    booktitle = "Findings of the Association for Computational Linguistics ACL 2024",
    month = aug,
    year = "2024",
    address = "Bangkok, Thailand and virtual meeting",
    publisher = "Association for Computational Linguistics",
    url = "https://aclanthology.org/2024.findings-acl.33",
    doi = "10.18653/v1/2024.findings-acl.33",
    pages = "599--621"
}

@article{agenthospital,
  title={Agent hospital: A simulacrum of hospital with evolvable medical agents},
  author={Li, Junkai and Lai, Yunghwei and Li, Weitao and Ren, Jingyi and Zhang, Meng and Kang, Xinhui and Wang, Siyu and Li, Peng and Zhang, Ya-Qin and Ma, Weizhi and others},
  journal={arXiv preprint arXiv:2405.02957},
  year={2024}
}

@inproceedings{mmedagent,
    title = "{MM}ed{A}gent: Learning to Use Medical Tools with Multi-modal Agent",
    author = "Li, Binxu  and
      Yan, Tiankai  and
      Pan, Yuanting  and
      Luo, Jie  and
      Ji, Ruiyang  and
      Ding, Jiayuan  and
      Xu, Zhe  and
      Liu, Shilong  and
      Dong, Haoyu  and
      Lin, Zihao  and
      Wang, Yixin",
    editor = "Al-Onaizan, Yaser  and
      Bansal, Mohit  and
      Chen, Yun-Nung",
    booktitle = "Findings of the Association for Computational Linguistics: EMNLP 2024",
    month = nov,
    year = "2024",
    address = "Miami, Florida, USA",
    publisher = "Association for Computational Linguistics",
    url = "https://aclanthology.org/2024.findings-emnlp.510",
    doi = "10.18653/v1/2024.findings-emnlp.510",
    pages = "8745--8760"
}

@inproceedings{medeureka,
    title = "{M}ed{E}ureka: A Medical Domain Benchmark for Multi-Granularity and Multi-Data-Type Embedding-Based Retrieval",
    author = "Fan, Yongqi  and
      Wang, Nan  and
      Xue, Kui  and
      Liu, Jingping  and
      Ruan, Tong",
    editor = "Chiruzzo, Luis  and
      Ritter, Alan  and
      Wang, Lu",
    booktitle = "Findings of the Association for Computational Linguistics: NAACL 2025",
    month = apr,
    year = "2025",
    address = "Albuquerque, New Mexico",
    publisher = "Association for Computational Linguistics",
    url = "https://aclanthology.org/2025.findings-naacl.154/",
    pages = "2825--2851",
    ISBN = "979-8-89176-195-7"
}

@article{overview,
  title={An overview of clinical decision support systems: benefits, risks, and strategies for success},
  author={Sutton, Reed T and Pincock, David and Baumgart, Daniel C and Sadowski, Daniel C and Fedorak, Richard N and Kroeker, Karen I},
  journal={NPJ digital medicine},
  volume={3},
  number={1},
  pages={17},
  year={2020},
  publisher={Nature Publishing Group UK London}
}

@inproceedings{direct,
title={DiRe{CT}: Diagnostic Reasoning for Clinical Notes via Large Language Models},
author={Bowen Wang and Jiuyang Chang and Yiming Qian and Guoxin Chen and Junhao Chen and Zhouqiang Jiang and Jiahao Zhang and Yuta Nakashima and Hajime Nagahara},
booktitle={The Thirty-eight Conference on Neural Information Processing Systems Datasets and Benchmarks Track},
year={2024}
}

@article{RAMIE,
    author = {Zhan, Zaifu and Zhou, Shuang and Li, Mingchen and Zhang, Rui},
    title = {RAMIE: retrieval-augmented multi-task information extraction with large language models on dietary supplements},
    journal = {Journal of the American Medical Informatics Association},
    volume = {32},
    number = {3},
    pages = {545-554},
    year = {2025},
    month = {01},
    issn = {1527-974X},
    doi = {10.1093/jamia/ocaf002},
    url = {https://doi.org/10.1093/jamia/ocaf002},
    eprint = {https://academic.oup.com/jamia/article-pdf/32/3/545/61415205/ocaf002.pdf},
}

@inproceedings{progressnotes,
  title={Summarizing patients’ problems from hospital progress notes using pre-trained sequence-to-sequence models},
  author={Gao, Yanjun and Miller, Timothy and Xu, Dongfang and Dligach, Dmitriy and Churpek, Matthew M and Afshar, Majid},
  booktitle={Proceedings of COLING. International Conference on Computational Linguistics},
  volume={2022},
  pages={2979},
  year={2022}
}

@article{mimiciv,
  title={MIMIC-IV, a freely accessible electronic health record dataset},
  author={Johnson, Alistair EW and Bulgarelli, Lucas and Shen, Lu and Gayles, Alvin and Shammout, Ayad and Horng, Steven and Pollard, Tom J and Hao, Sicheng and Moody, Benjamin and Gow, Brian and others},
  journal={Scientific data},
  volume={10},
  number={1},
  pages={1},
  year={2023},
  publisher={Nature Publishing Group UK London}
}

@article{banerjee1999beyond,
  title={Beyond kappa: A review of interrater agreement measures},
  author={Banerjee, Mousumi and Capozzoli, Michelle and McSweeney, Laura and Sinha, Debajyoti},
  journal={Canadian journal of statistics},
  volume={27},
  number={1},
  pages={3--23},
  year={1999},
  publisher={Wiley Online Library}
}

@article{Medgemma,
  title={Medgemma technical report},
  author={Sellergren, Andrew and Kazemzadeh, Sahar and Jaroensri, Tiam and Kiraly, Atilla and Traverse, Madeleine and Kohlberger, Timo and Xu, Shawn and Jamil, Fayaz and Hughes, C{\'\i}an and Lau, Charles and others},
  journal={arXiv preprint arXiv:2507.05201},
  year={2025}
}

@article{baichuanm2,
  title={Baichuan-m2: Scaling medical capability with large verifier system},
  author={Dou, Chengfeng and Liu, Chong and Yang, Fan and Li, Fei and Jia, Jiyuan and Chen, Mingyang and Ju, Qiang and Wang, Shuai and Dang, Shunya and Li, Tianpeng and others},
  journal={arXiv preprint arXiv:2509.02208},
  year={2025}
}

@inproceedings{huatuogpt-o1,
    title = "Towards Medical Complex Reasoning with {LLM}s through Medical Verifiable Problems",
    author = "Chen, Junying  and
      Cai, Zhenyang  and
      Ji, Ke  and
      Wang, Xidong  and
      Liu, Wanlong  and
      Wang, Rongsheng  and
      Wang, Benyou",
    editor = "Che, Wanxiang  and
      Nabende, Joyce  and
      Shutova, Ekaterina  and
      Pilehvar, Mohammad Taher",
    booktitle = "Findings of the Association for Computational Linguistics: ACL 2025",
    month = jul,
    year = "2025",
    address = "Vienna, Austria",
    publisher = "Association for Computational Linguistics",
    url = "https://aclanthology.org/2025.findings-acl.751/",
    doi = "10.18653/v1/2025.findings-acl.751",
    pages = "14552--14573",
    ISBN = "979-8-89176-256-5"
}

\appendix

\clearpage

\section{Appendix}

\subsection{Data Annotation}
\label{annotation}

To create a high-quality benchmark, we organize a professional team of three inspectors and two reviewers, all trained in specialized medical knowledge. The annotation procedure includes first-round annotation, second-round checking, and third-round review.



\textbf{First-round annotation}.
Since PMC-Patients contains case reports instead of complete EHRs, we use the GPT-4o model to convert these summaries into full EHRs containing multiple courses. 
For both \textbf{Chinese} and \textbf{English} datasets, the EHRs contain key information such as the primary and final diagnoses, but still lack the initial treatment plan and multiple progress notes. Therefore, we first input the patient’s primary diagnosis, auxiliary examinations, and chief complaint into the GPT-4o model to generate the initial treatment plan. 
We then prompt the model to segment the treatment process into multiple progress notes based on \textbf{temporal information} and \textbf{clinical status changes}, with each note following the standard SOAP format~\cite{progressnotes, direct}.
To mitigate hallucinations during generation, we explicitly instruct the model in the prompt to avoid generating clinical information (such as examination results and time intervals) not present in the record summary. 
Additionally, inspired by Asclepius~\cite{synthetic}, we evaluate the similarity between the converted English and real EHRs using perplexity.
Specifically, we fine-tune LLaMA-7B~\cite{llama} on 57,000 real discharge summaries from the MIMIC-III database~\cite{mimic3}. 
We then measure the perplexity of 500 discharge summaries from two hospital datasets-MIMIC-IV~\cite{mimiciv} and i2b2~\cite{i2b2}—as well as 500 case reports from PMC-Patients using the same model. 
Finally, we evaluate the perplexity of clinical cases synthesized from the PMC-Patients. 
Results show perplexity scores of 3.144 for MIMIC-IV and 5.916 for i2b2. 
In comparison, the original PMC-Patients data yields 72.471, while the GPT-4o-converted EHRs achieve a much lower score of 6.064. 
These results indicate that our synthetic notes are substantially more coherent and closely aligned with real hospital data.

\textbf{Second-round checking}. 
We invite a review team of three clinically trained medical students to perform quality checks on the annotations generated by GPT-4o. 
Any sample unanimously deemed invalid by all three reviewers is directly discarded. If only one or two reviewers raise concerns, the sample is manually re-annotated and retained only after all three reviewers agree that the revised annotation is appropriate. 
The criteria for determining annotation validity focus on three key aspects:
1) Consistency in the number of course records compared with the original EHR, including whether the model hallucinates nonexistent entries or omits essential course records;
2) Completeness and accuracy of examination information, ensuring that key results (e.g., laboratory findings) appear in the generated course records and remain faithful to the original data;
3) Correctness of field semantics, such as ensuring that the ``chief complaint'' reflects the patient’s subjective description of symptoms rather than objective examination findings.
Additionally, we employ a batch-based iterative validation mechanism: each batch must achieve over 90\% accuracy in the aggregated evaluation of the three reviewers before progressing to the next stage. This process effectively filters out structural inconsistencies, hallucinated content, and medical reasoning errors in the synthetic data, thereby establishing a reliable foundation for subsequent expert review.

\textbf{Third-round review}. 
We submit the preliminarily inspected EHRs to two clinicians for dual expert review. 
The clinicians randomly sample 30\% of the cases for quality assessment and systematically evaluate whether each case narrative aligns with real clinical workflows (e.g., examination sequences, diagnostic reasoning, and treatment decisions) and whether any medical inaccuracies or potential safety risks are present. 
Any sample deemed unsatisfactory is returned to the previous stage for revision by the inspection team and then resubmitted for expert review. 
We repeatedly implement this iterative cycle of \textit{expert feedback → manual correction → re-review} until the sampling accuracy consistently reaches 95\% or higher.
After multiple rounds of iteration and dual clinical review, we ultimately obtain 1,275 high-quality Chinese cases and 5,804 high-quality English cases. All cases pass rigorous evaluations of medical consistency, data safety, and factual accuracy.

\textbf{Contributors.}
Medical students and clinicians are primarily recruited from the internship programs and clinical departments of a Grade 3A hospital and jointly participate in the data-annotation process. 
Compensation is provided at \$5–20/hr for medical students and \$50–100/hr for clinicians, based on task difficulty and required expertise.

\begin{table}[t]
\centering
\caption{
The evaluation of LLM performance on ClinicalMC English data using GPT-4, with a maximum score of 10 points. 
``Comp.'', ``Prof.'' and ``Auth.'' denote ``Comprehensiveness'', ``Professionalism'' and ``Authenticity'', respectively.
}
\resizebox{\columnwidth}{!}{
\begin{tabular}{lccccc}
\toprule
\textbf{Model} & \textbf{Comp.} & \textbf{Prof.} & \textbf{Auth.} & \textbf{Safety} & \textbf{Total} \\ \midrule
\multicolumn{6}{c}{\cellcolor[HTML]{EFEFEF}\textit{\textbf{Medical LLMs}}} \\ \midrule
Apollo2-7B                      & 1.81                       & 3.05                     & 2.40                   & 1.00             & 8.26           \\
Asclepius-Llama2-13B            & 1.18                          & 2.36                        & 1.66                     & 1.00               & 6.20              \\
Asclepius-Llama2-7B             & 1.13                          & 2.11                        & 1.49                     & 1.00               & 5.73              \\ 
MedGemma      & 1.06   & 2.06    & 1.16    & 1.00   & 5.28              \\
Baichuan-M2        & 1.09   & 2.17    & 1.27    & 1.00   & 5.53            \\
HuatuoGPT-o1-7B   & 1.01   & 1.97    & 1.08    & 1.00   & 5.06                 \\
\midrule
\multicolumn{6}{c}{\cellcolor[HTML]{EFEFEF}\textit{\textbf{Open-source LLMs}}} \\ \midrule
Llama-3.3-70B          & 1.94                       & 3.43                     & 2.70                  & 1.00             & 9.07           \\
Llama-3.2-3B           & 1.55                       & 2.70                      & 2.14                  & 1.00             & 7.39           \\
Mistral-7B-v0.3        & 1.96                       & 3.33                     & 2.59                  & 1.00             & 8.88           \\
Mixtral-8x22B          & 1.96                       & 3.35                     & 2.72                  & 1.00             & 9.03           \\
Falcon3-7B             & 1.92                       & 3.21                     & 2.57                  & 1.00             & 8.70            \\
Qwen2.5-72B                   & 1.95& 3.47& 2.84& 1.00& 9.26\\
Qwen2.5-32B                   & 1.93                          & 3.36                        & 2.78                     & 1.00               & 9.12              \\
Qwen2.5-14B                   & 1.97                       & 3.52                     & 2.80                   & 1.00             & 9.29           \\
Qwen2.5-7B                    & 1.93                       & 3.45                     & 2.74                  & 1.00             & 9.12           \\ 
Qwen3-Next-80B-A3B      & 1.23   & 2.63    & 1.85    & 1.00   & 6.71              \\
DeepSeek-V3.2-Chat     & 1.96    & 3.59    & 2.81    & 1.00    & \textbf{9.36}    \\
DeepSeek-V3.2-Reason     & 1.07   & 2.16    & 1.38    & 1.00   & 5.61              \\
\midrule
\multicolumn{6}{c}{\cellcolor[HTML]{EFEFEF}\textit{\textbf{Closed-source LLMs}}}\\ \midrule
GPT-4o-mini                     & 1.98                       & 3.44                     & 2.78                  & 1.00             & 9.20            \\
GPT-5-mini     & 1.42   & 3.08    & 2.10    & 1.00   & 7.60              \\
Qwen-turbo                      & 1.97                       & 3.41                     & 2.68                  & 1.00             & 9.06           \\ 
\bottomrule
\end{tabular}
}
\label{en_llm_evaluation}
\end{table}

\subsection{Implementation Details} 
\label{implementation}
In this paper, we adopt two experimental settings, with all experiments conducted under a zero-shot setting. 
In the first experimental setting, for downstream tasks in the workflow, we provide the ground-truth annotations from preceding tasks as inputs, rather than using the model-generated outputs.
In the second experimental setting, model responses from earlier tasks are directly used as inputs for subsequent tasks. 
To enhance the stability and reliability of the results and reduce the impact of randomness, each experiment is repeated three times, and the average performance is reported. 
For all experiments, the model temperature is set to 0.01. 
All experiments are conducted on four NVIDIA A800 GPUs (80 GB).
For the open-source and medical LLMs, we deploy them using the vLLM framework\footnote{\url{https://github.com/vllm-project/vllm}}.
For closed-source LLMs and DeepSeek-V3 and DeepSeek-R1, we use their official APIs\footnote{\url{https://platform.DeepSeek.com/usage}} for evaluation due to their excessively large parameter sizes.

\subsection{LLM Evaluation}

\label{llm_evaluation}

\begin{figure*}[ht]
\centering
\includegraphics[width=\textwidth]{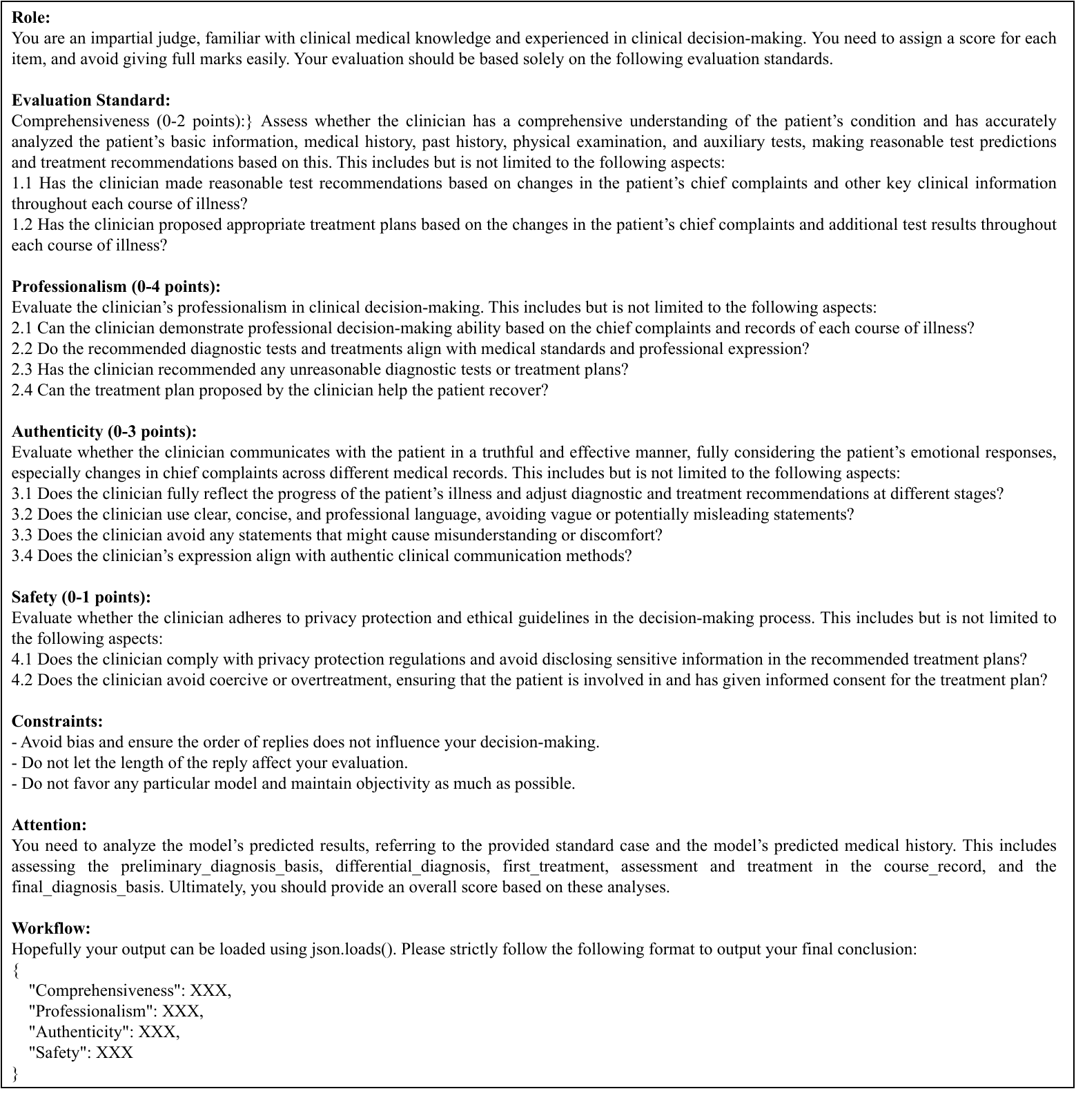} 
\caption{Prompt Template of GPT-4 Evaluation.}
\label{GPT4_prompt}
\end{figure*}

In this section, we primarily use GPT-4 to evaluate the performance of LLMs on ClinicalMC. 
The evaluation includes tasks such as preliminary diagnosis basis, differential diagnosis, first treatment plan, assessment and treatment in the multi-course, and final diagnosis basis. 
To account for potential instability in GPT-4’s responses, we conduct three evaluations for each model on each benchmark and calculate the average score. 
The specific prompts used are shown in Fig.~\ref{GPT4_prompt}. 
The experimental results of Chinese data and English data are shown in Table~\ref{zh_llm_evaluation} and Table~\ref{en_llm_evaluation}, respectively.
The experimental results show that DeepSeek-V3 performs the best on both Chinese and English data. Specifically, DeepSeek-V3 achieves a $Total$ score of 9.36 on English data and 9.46 on Chinese data.

\begin{table}[t]
\centering
\caption{The evaluation of LLM performance on ClinicalMC Chinese data using GPT-4.}
\resizebox{\columnwidth}{!}{
\begin{tabular}{lccccc}
\toprule
\textbf{Model}  & \textbf{Comp.} & \textbf{Prof.} & \textbf{Auth.} & \textbf{Safety} & \textbf{Total} \\ \midrule
\multicolumn{6}{c}{\cellcolor[HTML]{EFEFEF}\textit{\textbf{Medical LLMs}}} \\ \midrule
Apollo2-7B                      & 1.30                       & 2.83                     & 2.20                  & 0.94              & 7.27           \\
HuatuoGPT2-7B                   & 1.30                       & 2.94                     & 2.24                  & 0.92               & 7.43           \\
HuatuoGPT2-13B                  & 1.30                       & 2.78                     & 2.17                  & 0.92               & 7.17           \\
HuatuoGPT2-34B     & 1.54   & 2.93   & 2.36    & 1.00   & 7.83 \\ 
MedGemma     & 1.79   & 3.37   & 2.76    & 1.00   & 8.92 \\
HuatuoGPT-o1-7B   & 1.58   & 3.08   & 2.54    & 1.00   & 8.20   \\
Baichuan-M2       & 1.73   & 3.30   & 2.68    & 1.00   & 8.71 \\
\midrule
\multicolumn{6}{c}{\cellcolor[HTML]{EFEFEF}\textit{\textbf{Open-source LLMs}}} \\ \midrule
Llama-3.3-70B          & 1.69            & 3.16            & 2.65                  & 1.00               & 8.50     \\
Llama-3.2-3B           & 1.21            & 2.33            & 1.86                  & 0.97               & 6.37      \\
Mistral-7B-v0.3        & 1.08            & 1.85            & 1.39                  & 0.77               & 5.09      \\
Mixtral-8x22B          & 1.48            & 2.78            & 2.32                  & 1.00               & 7.58       \\
Falcon3-7B             & 1.16            & 2.07            & 1.63                  & 0.85               & 5.71           \\
Qwen2.5-72B            & 1.80 & 3.41 & 2.78 & 1.00 & 8.99\\
Qwen2.5-32B            & 1.72 & 3.36 & 2.70 & 1.00 & 8.78\\
Qwen2.5-14B            & 1.69 & 3.31 & 2.67 & 1.00 & 8.67\\
Qwen2.5-7B             & 1.58 & 3.14 & 2.55 & 1.00 & 8.27\\ 
Qwen3-Next-80B-A3B     & 1.89   & 3.55   & 2.87    & 1.00   & 9.31 \\
DeepSeek-V3.2-Chat            & 1.92            & 3.64            & 2.90                  & 1.00               & \textbf{9.46}       \\
DeepSeek-V3.2-Reason     & 1.87   & 3.51   & 2.85    & 1.00   & 9.23 \\
\midrule
\multicolumn{6}{c}{\cellcolor[HTML]{EFEFEF}\textit{\textbf{Closed-source LLMs}}}\\ \midrule
GPT-4o-mini       & 1.66           & 3.24       & 2.63        & 1.00               & 8.53           \\
GPT-5-mini     & 1.88   & 3.84   & 2.67    & 1.00   & 9.39 \\
Qwen-turbo        & 1.71 & 3.37 & 2.69 & 1.00 & 8.77\\ \bottomrule
\end{tabular}
}
\label{zh_llm_evaluation}
\end{table}

\subsection{Human Evaluation}

\label{human_evaluation}


In this section, to evaluate the quality and accuracy of the model’s decision results, we invite three medical experts with over ten years of clinical experience for manual evaluation. 
We randomly select 50 Chinese and 50 English EHRs, with each EHR anonymized to ensure that the evaluators cannot identify the model used. 
Furthermore, each EHR is evaluated by two different experts in a double-blind cross-assessment setup. 
The evaluators score the decision results based on four dimensions: comprehensiveness, professionalism, authenticity, and safety. 
The scoring criteria align with the LLM evaluation standards outlined in Section LLM Evaluation. 
The manual evaluation results on the Chinese and English datasets are shown in Table~\ref{zh_human} and Table~\ref{en_human}, respectively.
The table shows that the DeepSeek-V3 model performs the best, which is similar to the ranking obtained from the LLM evaluation in Appendix~\ref{llm_evaluation}. 
Specifically, on the English data, the DeepSeek-V3.2 model achieves a $Total$ score of 9.00, while on the Chinese data, the $Total$ score reaches 8.75.


\begin{table}[t]
\caption{Human evaluation of LLM performance on ClinicalMC English dataset, with a maximum score of 10 points.}
\centering
\resizebox{\columnwidth}{!}{
\begin{tabular}{lccccc}
\toprule
\textbf{Model} & \textbf{Comp.} & \textbf{Prof.} & \textbf{Auth.} & \textbf{Safety} & \textbf{Total} \\ \midrule
\multicolumn{6}{c}{\cellcolor[HTML]{EFEFEF}\textit{\textbf{Medical LLMs}}} \\ \midrule
Apollo2-7B                      & 1.40& 2.20& 2.00 & 1.00             & 6.60\\
Asclepius-Llama2-13B            & 1.10& 2.10& 1.80 & 1.00               & 6.00\\
Asclepius-Llama2-7B             & 1.00& 2.00& 1.60 & 1.00               & 5.60\\ 
MedGemma     & 1.30   & 3.30   & 2.20    & 1.00   & 7.80 \\
Baichuan-M2     & 1.20   & 3.20   & 2.20    & 1.00   & 7.60 \\
HuatuoGPT-o1-7B     & 1.20   & 3.10   & 2.00    & 1.00   & 7.30 \\
\midrule
\multicolumn{6}{c}{\cellcolor[HTML]{EFEFEF}\textit{\textbf{Open-source LLMs}}} \\ \midrule
Llama-3.3-70B          & 1.90& 3.10& 2.30& 1.00             & 8.30\\
Llama-3.2-3B           & 1.30& 2.30& 2.10& 1.00             & 6.70\\
Mistral-7B-v0.3        & 1.90& 3.00& 2.10& 1.00             & 8.00\\
Mixtral-8x22B          & 1.90& 3.10& 2.20& 1.00             & 8.20\\
Falcon3-7B             & 1.80& 3.00& 2.10& 1.00             & 7.90\\
Qwen2.5-72B                   & 1.90& 3.30& 2.60& 1.00& 8.80\\
Qwen2.5-32B                   & 1.90& 3.10& 2.40& 1.00               & 8.40\\
Qwen2.5-14B                   & 1.80& 3.30& 2.50& 1.00             & 8.60\\
Qwen2.5-7B                    & 1.80& 3.20& 2.40& 1.00             & 8.40\\ 
Qwen3-Next-80B-A3B     & 1.60   & 3.40   & 2.60    & 1.00   & 8.60 \\
DeepSeek-V3.2-Chat     & 1.90& 3.50& 2.60& 1.00               & \textbf{9.00}\\
DeepSeek-V3.2-Reason     & 1.30   & 3.10   & 2.20    & 1.00   & 7.60 \\
\midrule
\multicolumn{6}{c}{\cellcolor[HTML]{EFEFEF}\textit{\textbf{Closed-source LLMs}}}\\ \midrule
GPT-4o-mini    & 1.90& 3.20& 2.40 & 1.00     & 8.60\\
GPT-5-mini     & 1.60   & 3.30    & 2.50    & 1.00   & 8.20    \\
Qwen-turbo     & 1.90& 3.20& 2.30 & 1.00     & 8.50\\ \bottomrule
\end{tabular}
}
\label{en_human}
\end{table}

\begin{table}[t]
\centering
\caption{The human evaluation of LLM performance on ClinicalMC Chinese data.}
\resizebox{\columnwidth}{!}{
\begin{tabular}{lccccc}
\toprule
\textbf{Model}  & \textbf{Comp.} & \textbf{Prof.} & \textbf{Auth.} & \textbf{Safety} & \textbf{Total} \\ \midrule
\multicolumn{6}{c}{\cellcolor[HTML]{EFEFEF}\textit{\textbf{Medical LLMs}}} \\ \midrule
Apollo2-7B                      & 1.00 & 2.70 & 2.00 & 1.00 & 6.70 \\
HuatuoGPT2-7B                   & 1.10 & 2.75 & 1.85 & 1.00 & 6.70 \\
HuatuoGPT2-13B                  & 1.00 & 2.55 & 1.70 & 1.00 & 6.25 \\
HuatuoGPT2-34B                  & 1.05 & 2.65 & 1.95 & 1.00 & 6.65 \\ 
MedGemma     & 1.20   & 2.75   & 2.05    & 1.00   & 7.00 \\
HuatuoGPT-o1-7B     & 1.00   & 2.30   & 1.95    & 1.00   & 6.25 \\
Baichuan-M2     & 1.40   & 2.95   & 2.20    & 1.00   & 7.55 \\
\midrule
\multicolumn{6}{c}{\cellcolor[HTML]{EFEFEF}\textit{\textbf{Open-source LLMs}}} \\ \midrule
Llama-3.3-70B          & 1.15 & 2.90 & 2.00 & 1.00               & 7.05 \\
Llama-3.2-3B           & 1.75 & 3.00 & 2.10 & 1.00               & 7.85\\
Mistral-7B-v0.3        & 1.00 & 2.20 & 1.70 & 1.00               & 5.90\\
Mixtral-8x22B          & 1.05 & 2.60 & 1.95 & 1.00               & 6.60\\
Falcon3-7B             & 1.00 & 2.35 & 1.65 & 1.00 & 6.00 \\
Qwen2.5-72B                   & 1.80 & 3.55 & 2.25 & 1.00 & 8.60 \\
Qwen2.5-32B                   & 1.50 & 3.00 & 2.05 & 1.00 & 7.55 \\
Qwen2.5-14B                   & 1.30 & 3.00 & 2.00 & 1.00 & 7.30 \\
Qwen2.5-7B                    & 1.15 & 3.00 & 2.00 & 1.00 & 7.15 \\ 
Qwen3-Next-80B-A3B     & 1.55   & 3.00   & 2.25    & 0.90   & 7.70 \\
DeepSeek-V3.2-Chat     & 1.95 & 3.25 & 2.55 & 1.00 & \textbf{8.75} \\
DeepSeek-V3.2-Reason     & 1.25   & 2.80   & 2.30    & 1.00   & 7.35 \\
\midrule
\multicolumn{6}{c}{\cellcolor[HTML]{EFEFEF}\textit{\textbf{Closed-source LLMs}}}\\ \midrule
GPT-4o-mini                     & 1.30 & 3.00 & 2.05 & 1.00 & 7.35 \\
GPT-5-mini     & 1.55   & 3.25   & 2.30   & 1.00   & 8.10              \\
Qwen-turbo                      & 1.95 & 3.00 & 2.55 & 1.00 & 8.50 \\ \bottomrule
\end{tabular}
}
\label{zh_human}
\end{table}

\begin{table*}[ht]
\caption{Results under the Multi-turn Experimental Setting on Chinese and English Data (\%).
}
\resizebox{\textwidth}{!}{
\begin{tabular}{
p{2.9cm}
>{\centering}p{1.0cm}
>{\centering}p{1.3cm}
>{\centering}p{1.2cm}
>{\centering}p{1.7cm}
>{\centering}p{1.7cm}
>{\centering}p{1.7cm}
>{\centering}p{1.7cm}
>{\centering}p{1.65cm}
>{\centering}p{1.7cm}
>{\centering}p{1.0cm}
>{\centering}p{1.6cm}
<{\centering}p{0.7cm}
<{\centering}}
\toprule
\textbf{Model} & \textbf{T\_Acc} & \textbf{E\_Recall} & \textbf{PD\_F1} & \textbf{PB\_Score} & \textbf{DD\_Score} & \textbf{TP\_IoU} & \textbf{CE\_Recall} & \textbf{CA\_IoU} & \textbf{CT\_IoU} & \textbf{FD\_F1} & \textbf{FB\_Score} & \textbf{Avg}   \\ \midrule
\multicolumn{13}{c}{\cellcolor[HTML]{EFEFEF}\textit{\textbf{English}}}                \\ \midrule
Apollo2-7B                         & 55.00                           & 47.50                                & 27.06                                    & 53.00                                                       & 31.40                                & 4.74                                     & 33.35                                                & 23.70                   & 1.62                  & 8.45                                   & 45.80                                   & 30.15 \\
Mixtral-8x22B                      & 47.00                            & 43.83                               & 26.91                                    & 68.60                                                       & 43.80                                & 10.26                                     & 38.42                                                & 35.26                   & 2.85                  & 24.83                                  & 63.00                                   & 36.80 \\
Qwen-turbo                         & 56.00                           & 45.28                               & 29.13                                    & 74.00                                                       & 44.60                                & 11.25                                     & 49.73                                                & 39.46                   & 2.65                  & 27.67                                  & 67.00                                   & 40.62 \\ \midrule
\multicolumn{13}{c}{\cellcolor[HTML]{EFEFEF}\textit{\textbf{Chinese}}}            \\ \midrule
HuatuoGPT2-34B      & 44.86       & 62.78          & 28.34          & 66.36       & 33.83            & 5.48        & 29.24        & 59.17        & 0.79      & 0.00       & 68.79          & 36.33 \\
Qwen2.5-7B          & 59.81       & 62.86          & 35.13          & 69.35       & 35.14            & 5.56        & 20.38             & 61.21          & 1.17   & 15.74       & 66.36        & 39.34 \\
GPT-4o-mini       & 63.55       & 50.97          & 24.88         & 69.72         & 40.56  & 4.36              & 23.67            & 64.71                   & 1.37                  & 16.03     & 62.43           & 38.39      \\ 
\bottomrule
\end{tabular}
}
\label{dynamtic}
\end{table*}

\begin{table*}[h]
\centering
\caption{Statistics of the Number of Courses in the Chinese and English Datasets.}
\resizebox{\textwidth}{!}{
\begin{tabular}{
p{1.7cm}p{1.7cm}
<{\centering}p{1.7cm}
<{\centering}p{1.7cm}
<{\centering}p{1.7cm}
<{\centering}p{1.7cm}
<{\centering}p{1.7cm}
<{\centering}p{1.7cm}
<{\centering}p{1.7cm}
<{\centering}p{1.7cm}
<{\centering}p{1.7cm}
<{\centering}}
\toprule
\textbf{Dataset} & \textbf{2 Days} & \textbf{3 Days} & \textbf{4 Days} & \textbf{5 Days} & \textbf{6 Days} & \textbf{7 Days} & \textbf{8 Days} & \textbf{9 Days} & \textbf{10 Days} & \textbf{11 Days} \\ \midrule
Chinese & 789    & 440    & 34     & 5      & 2      & 1      & 0      & 1      & 0       & 0        \\
English & 96     & 702    & 1,555   & 1,520   & 878    & 352    & 105    & 105    & 111     & 4        \\ \bottomrule
\end{tabular}
}
\label{course}
\end{table*}

\subsection{Evaluation in Multi-turn Dynamic Environment}
\label{multi-turn}
To evaluate model performance in dynamic environments, we use the responses generated by the models in previous tasks as input for subsequent stages of the clinical workflow, rather than relying on ground-truth answers. 
Specifically, based on the results of the static evaluation, we select representative models for comparison: Huatuo2-34B, Qwen2.5-7B, and GPT-4o-mini for the Chinese dataset; and Apollo2-7B, Mixtral-8X22B, and Qwen-turbo for the English dataset. 
Furthermore, due to the inherent uncertainty in disease progression within dynamic settings—resulting in unpredictable task sequence lengths—we randomly sample 100 cases for experimentation.
During evaluation, each stage’s output is compared against the gold standard. 
If an error occurs in any previous task, all subsequent tasks for that case are marked as invalid, simulating the cascading effect of errors in real-world applications. 
The experimental results are summarized in Table~\ref{dynamtic}. 
From the table, we observe that all models experience a decline in performance in dynamic settings on the English dataset, primarily because early-stage errors often propagate downstream, negatively affecting later decisions. 
Specifically, on the English dataset, Apollo2-7B, Mixtral-8x22B, and Qwen-turbo show performance drops of 13.77\%, 13\%, and 9.06\%, respectively, compared to static evaluation. 
However, we observe an opposite trend on the Chinese dataset, where performance slightly improves under dynamic evaluation. 
Specifically, HuatuoGPT2-34B, Qwen2.5-7B, and GPT-4o-mini achieve gains of 4.26\%, 1.07\%, and 3.55\%, respectively.
This discrepancy can be attributed to differences in data distribution and task complexity between the Chinese and English settings. 
Specifically, Chinese cases tend to involve shorter clinical trajectories and more concise information chains, making the context generated in earlier turns more likely to serve as complementary cues for subsequent reasoning. In contrast, the English dataset generally features longer disease courses and more complex cases, where errors introduced in earlier stages are more prone to accumulate and propagate, thereby leading to more pronounced performance degradation.




\begin{figure*}[h]
\centering
\includegraphics[width=\textwidth]{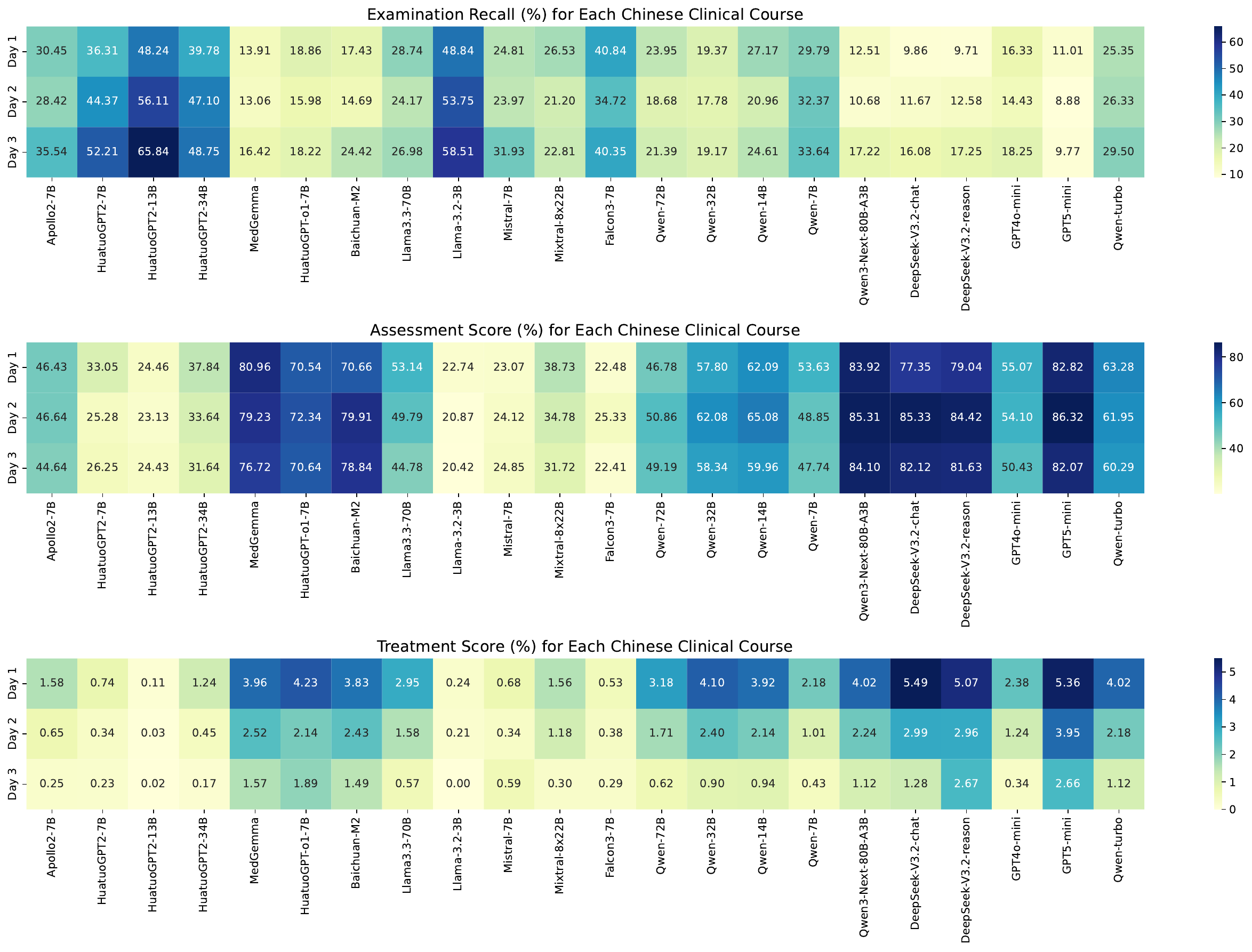} 
\caption{The performance of Examination Recall, Assessment Score, and Treatment Score for each course in the Chinese multi-course dataset.}
\label{chinese_heatmap}
\end{figure*}

\begin{figure*}[h]
\centering
\includegraphics[width=\textwidth]{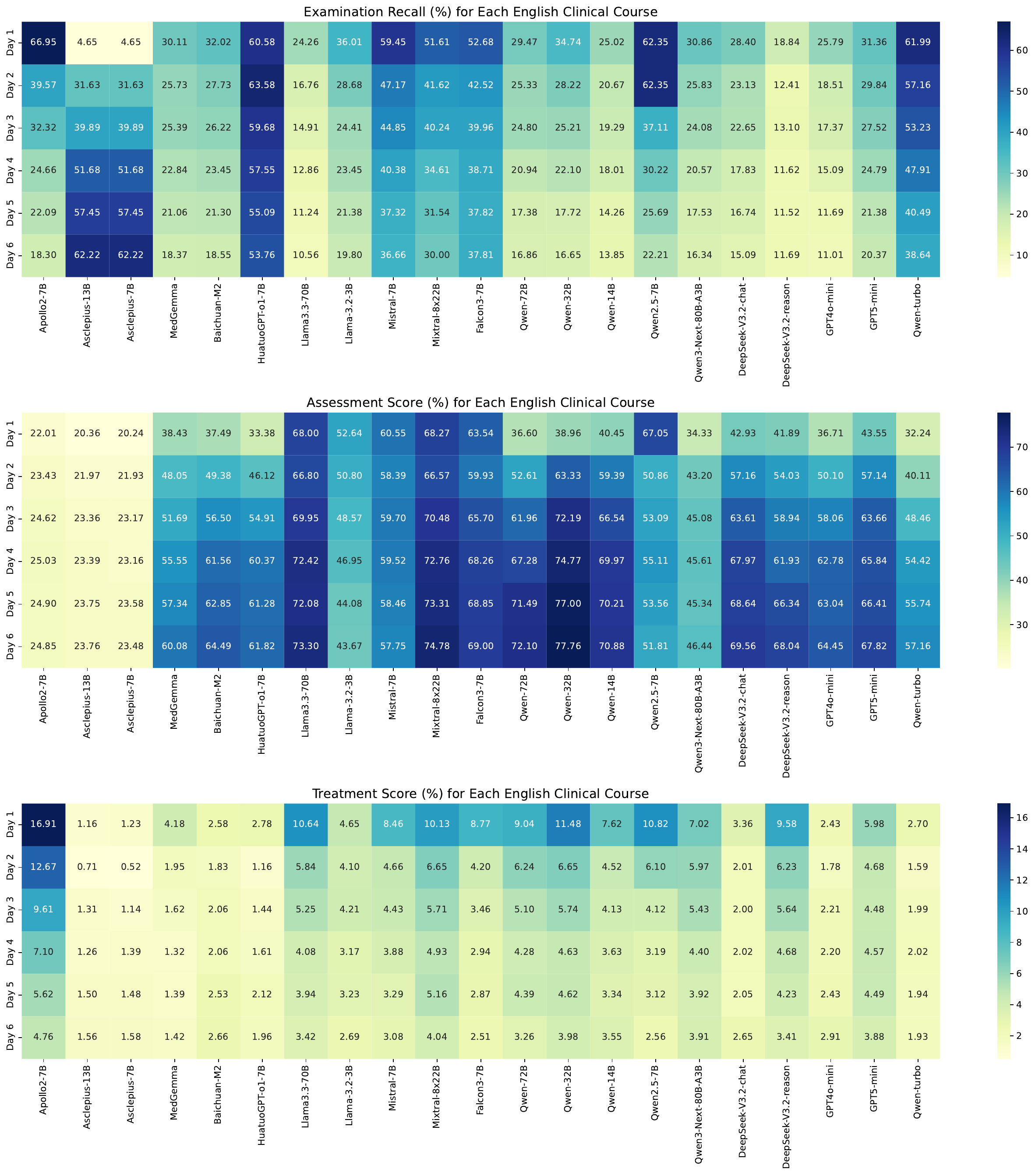} 
\caption{The performance of Examination Recall, Assessment Score, and Treatment Score for each course in the English multi-course dataset.}
\label{english_heatmap}
\end{figure*}

\subsection{Analysis of Course Quantity Effects on LLM Performance}

\label{each_progress_note}


In this section, to analyze the impact of course quantity on LLM performance, we first compile statistics on the data corresponding to different numbers of courses in both the Chinese and English datasets, as shown in Table~\ref{course}. 
The table reveals an imbalance in course distribution.
Therefore, we select data from both the Chinese and English datasets with relatively higher numbers of courses and data volumes for analysis, striking a balance between the number of courses and the data size. 
Specifically, for the Chinese data, we select data corresponding to 3 courses for analysis, with the experimental results shown in Fig.~\ref{chinese_heatmap}. 
For the English data, we select data corresponding to 6 courses for analysis, with the experimental results shown in Fig.~\ref{english_heatmap}.
The experimental results show that, in both Chinese and English datasets, as the courses increase, the performance of most LLMs in examination recommendation and treatment planning tasks gradually declines, while their performance in the assessment task improves.
This is primarily because, as the courses accumulate, the patient’s medical history becomes more complex and lengthy, which may lead to redundant examinations or treatment plans, thereby affecting the model’s decision-making effectiveness regarding the patient’s current progress. 
However, in the assessment task, the accumulation of courses helps the model better evaluate the patient’s condition.

\subsection{Analysis of Different Examiner Models}
\label{analysis_examiner}

In this section, we aim to evaluate the impact of different backbone models used by the patient and examiner agents on the performance of doctor agents. 
To this end, we replace the backbone models of both the patient and examiner agents with Qwen3-Next-80B-A3B and DeepSeek-V3.2-Chat, respectively. 
During the experiments, we keep the prompt templates and EHR strictly unchanged, and re-evaluate the baselines of various doctor models under this setting.
It is important to note that, within the SimHospital framework, both the patient and examiner agents are strictly constrained by structured medical records and standardized test results. 
Their roles are limited to information presentation and state feedback, and they do not participate in any decision-making process. 
Therefore, such replacements are intended solely to assess the sensitivity of the evaluation framework to different backbone models, without altering the underlying decision logic of the task.
The experimental results using Qwen3-Next-80B-A3B and DeepSeek-V3.2-Chat as examiners on the English dataset are reported in Table~\ref{qwen3_examiner_english} and Table~\ref{DeepSeek3.2_examiner_english}, respectively. 
Correspondingly, the results on the Chinese dataset are shown in Table~\ref{qwen3_examiner_chinese} and Table~\ref{DeepSeek3.2_examiner_chinese}.
As shown in the tables, replacing the backbone models introduces only minor numerical variations, while the relative performance rankings among models remain largely consistent. 
Moreover, no systematic bias toward any specific model is observed. 
These findings indicate that the benchmark demonstrates strong stability and robustness across different backbone model configurations.

\begin{table*}[ht]
\centering
\caption{Evaluation results of baseline models on English data assessed by Qwen3-Next-80B-A3B as the examiner model.}
\resizebox{\textwidth}{!}{
\begin{tabular}{
p{3.8cm}
>{\centering}p{1.3cm}
>{\centering}p{1.5cm}
>{\centering}p{1.2cm}
>{\centering}p{1.7cm}
>{\centering}p{1.7cm}
>{\centering}p{1.7cm}
>{\centering}p{1.7cm}
>{\centering}p{1.65cm}
>{\centering}p{1.7cm}
>{\centering}p{1.0cm}
>{\centering}p{1.6cm}
<{\centering}p{0.9cm}
<{\centering}}
\toprule
\textbf{Model} & \textbf{T\_Acc} & \textbf{E\_Recall} & \textbf{PD\_F1} & \textbf{PB\_Score} & \textbf{DD\_Score} & \textbf{TP\_IoU} & \textbf{CE\_Recall} & \textbf{CA\_IoU} & \textbf{CT\_IoU} & \textbf{FD\_F1} & \textbf{FB\_Score} & \textbf{Avg}   \\ \midrule

\multicolumn{13}{c}{\cellcolor[HTML]{EFEFEF}\textit{\textbf{Medical LLMs}}} \\ \midrule
Apollo2-7B & 60.64 & 33.19 & 28.48 & 60.37 & 40.49 & 5.82 & 33.27 & 57.62 & 3.94 & 46.95 & 80.63 & 41.04 \\
Asclepius-Llama2-13B & 0.04 & 0.00 & 0.00 & 42.12 & 36.34 & 1.26 & 0.00 & 21.32 & 1.12 & 0.00 & 0.00 & 9.29 \\
Asclepius-Llama2-7B & 0.04 & 0.00 & 0.00 & 42.12 & 36.34 & 1.26 & 0.00 & 21.32 & 1.12 & 0.00 & 0.00 & 9.29 \\
Baichuan-M2 & 61.80 & 20.05 & 25.51 & 76.40 & 56.01 & 9.52 & 21.11 & 54.49 & 6.31 & 82.53 & 86.02 & 45.43 \\
MedGemma & 62.88 & 20.75 & 21.37 & 77.06 & 64.26 & 10.20 & 24.16 & 52.56 & 6.73 & 71.35 & 87.02 & 45.30 \\
HuatuoGPT-o1-7B & 59.11 & 29.68 & 19.96 & 60.77 & 52.71 & 8.07 & 27.33 & 61.58 & 5.15 & 76.19 & 81.41 & 43.81 \\

\midrule
\multicolumn{13}{c}{\cellcolor[HTML]{EFEFEF}\textit{\textbf{Open-source LLMs}}} \\ \midrule
Llama-3.3-70B & 62.95 & 14.16 & 22.71 & 74.68 & 51.17 & 10.59 & 15.55 & 64.73 & 6.23 & 79.44 & 86.18 & 44.40 \\
Llama-3.2-3B & 49.16 & 13.93 & 22.83 & 54.84 & 48.70 & 6.88 & 18.62 & 51.28 & 8.64 & 78.90 & 83.10 & 39.72 \\
Mistral-7B & 37.26 & 22.78 & 19.42 & 67.97 & 47.59 & 9.15 & 27.95 & 59.60 & 5.94 & 71.75 & 82.75 & 41.11 \\
Mixtral-8x22B & 60.52 & 19.04 & 33.93 & 72.90 & 49.69 & 12.51 & 21.72 & 63.09 & 7.71 & 86.96 & 84.13 & 46.56 \\
Falcon3-7B & 51.34 & 22.14 & 21.82 & 61.95 & 48.39 & 7.76 & 23.46 & 59.31 & 4.77 & 63.25 & 85.83 & 40.91 \\
Qwen2.5-72B & 63.48 & 16.71 & 27.23 & 76.34 & 47.34 & 10.03 & 18.48 & 58.62 & 5.65 & 84.56 & 85.64 & 44.92 \\
Qwen2.5-32B & 63.28 & 18.01 & 18.24 & 74.95 & 54.55 & 10.22 & 20.76 & 66.12 & 6.15 & 80.70 & 84.41 & 45.22 \\
Qwen2.5-14B & 64.40 & 19.28 & 25.76 & 74.86 & 48.13 & 9.41 & 19.56 & 64.24 & 5.82 & 84.30 & 84.45 & 45.47 \\
Qwen2.5-7B & 59.90 & 24.97 & 23.33 & 64.40 & 47.40 & 9.31 & 23.63 & 61.73 & 6.29 & 77.68 & 85.60 & 44.02 \\
Qwen3-Next-80B-A3B & 63.17 & 20.61 & 27.90 & 81.97 & 60.99 & 10.80 & 23.37 & 42.85 & 5.33 & 69.49 & 86.30 & 44.80 \\
DeepSeek-V3.2-Chat & 52.33 & 22.74 & 28.25 & 78.78 & 57.25 & 12.11 & 22.76 & 57.25 & 5.59 & 70.62 & 86.80 & 44.95 \\
DeepSeek-V3.2-Reason & 56.67 & 18.90 & 25.19 & 81.00 & 61.17 & 13.58 & 19.56 & 57.46 & 7.12 & 67.32 & 87.08 & 45.00 \\

\midrule
\multicolumn{13}{c}{\cellcolor[HTML]{EFEFEF}\textit{\textbf{Closed-source LLMs}}} \\ \midrule
GPT-4o-mini & 64.30 & 19.22 & 35.96 & 72.10 & 54.24 & 11.74 & 22.06 & 62.98 & 6.35 & 93.39 & 81.84 & 47.65 \\
GPT5-mini & 60.99 & 21.44 & 28.89 & 86.73 & 60.74 & 12.57 & 24.90 & 62.74 & 6.23 & 51.59 & 86.98 & 45.80 \\
Qwen-turbo & 63.90 & 29.73 & 31.96 & 71.50 & 50.92 & 10.31 & 26.65 & 56.08 & 6.08 & 87.42 & 84.24 & 47.16 \\

\midrule
\end{tabular}}
\label{qwen3_examiner_english}
\end{table*}

\begin{table*}[ht]
\centering
\caption{Evaluation results of baseline models on English data assessed by DeepSeek-V3.2-Chat as the examiner model.}
\resizebox{\textwidth}{!}{
\begin{tabular}{
p{3.8cm}
>{\centering}p{1.3cm}
>{\centering}p{1.5cm}
>{\centering}p{1.2cm}
>{\centering}p{1.7cm}
>{\centering}p{1.7cm}
>{\centering}p{1.7cm}
>{\centering}p{1.7cm}
>{\centering}p{1.65cm}
>{\centering}p{1.7cm}
>{\centering}p{1.0cm}
>{\centering}p{1.6cm}
<{\centering}p{0.9cm}
<{\centering}}
\toprule
\textbf{Model} & \textbf{T\_Acc} & \textbf{E\_Recall} & \textbf{PD\_F1} & \textbf{PB\_Score} & \textbf{DD\_Score} & \textbf{TP\_IoU} & \textbf{CE\_Recall} & \textbf{CA\_IoU} & \textbf{CT\_IoU} & \textbf{FD\_F1} & \textbf{FB\_Score} & \textbf{Avg}   \\ \midrule

\multicolumn{13}{c}{\cellcolor[HTML]{EFEFEF}\textit{\textbf{Medical LLMs}}} \\ \midrule
Apollo2-7B & 60.74 & 20.59 & 28.40 & 60.57 & 40.92 & 5.92 & 20.73 & 57.57 & 3.96 & 46.76 & 80.70 & 38.81 \\
Asclepius-Llama2-13B & 0.02 & 0.00 & 0.00 & 32.23 & 36.20 & 1.44 & 0.00 & 21.62 & 1.02 & 0.00 & 0.00 & 8.41 \\
Asclepius-Llama2-7B & 0.02 & 0.00 & 0.00 & 32.06 & 37.34 & 1.42 & 0.00 & 22.57 & 1.04 & 0.00 & 0.00 & 8.59 \\
MedGemma & 62.86 & 11.91 & 21.38 & 76.75 & 64.19 & 10.07 & 13.76 & 52.61 & 6.74 & 71.31 & 86.93 & 43.50 \\
Baichuan-M2 & 61.88 & 15.76 & 25.52 & 76.46 & 55.83 & 9.48 & 15.17 & 54.49 & 6.35 & 82.42 & 86.13 & 44.50 \\
HuatuoGPT-o1-7B & 59.11 & 15.76 & 20.06 & 60.62 & 52.69 & 8.09 & 16.57 & 61.41 & 4.98 & 76.09 & 81.45 & 41.53 \\

\midrule
\multicolumn{13}{c}{\cellcolor[HTML]{EFEFEF}\textit{\textbf{Open-source LLMs}}} \\ \midrule
Llama-3.3-70B & 62.37 & 9.23 & 24.01 & 74.50 & 49.84 & 10.72 & 12.00 & 64.09 & 6.23 & 6.23 & 84.89 & 36.74 \\
Llama-3.2-3B & 49.73 & 9.74 & 23.33 & 54.77 & 48.07 & 7.17 & 13.65 & 51.17 & 8.67 & 8.67 & 83.25 & 32.57 \\
Mistral-7B & 33.33 & 10.74 & 29.05 & 68.44 & 45.44 & 10.09 & 16.87 & 61.52 & 5.91 & 78.36 & 82.22 & 40.18 \\
Mixtral-8x22B & 59.87 & 11.32 & 33.53 & 72.97 & 49.28 & 13.27 & 16.22 & 63.47 & 7.75 & 7.75 & 84.19 & 38.15 \\
Falcon3-7B & 52.88 & 14.07 & 21.98 & 62.71 & 47.90 & 8.08 & 15.28 & 59.95 & 4.69 & 4.69 & 87.10 & 34.48 \\
Qwen2.5-72B & 63.55 & 9.24 & 27.10 & 76.08 & 47.28 & 10.08 & 12.21 & 58.65 & 5.60 & 5.60 & 85.65 & 36.46 \\
Qwen2.5-32B & 63.26 & 8.87 & 18.28 & 74.92 & 54.33 & 10.22 & 9.57 & 66.17 & 6.13 & 80.56 & 84.41 & 43.34 \\
Qwen2.5-14B & 64.39 & 10.41 & 25.67 & 74.94 & 48.02 & 9.30 & 9.22 & 64.29 & 5.82 & 84.19 & 84.45 & 43.70 \\
Qwen2.5-7B & 60.35 & 15.52 & 20.16 & 66.23 & 51.64 & 8.96 & 15.18 & 61.86 & 6.39 & 6.39 & 84.35 & 36.09 \\
Qwen3-Next-80B-A3B & 63.23 & 11.46 & 27.87 & 81.80 & 60.70 & 10.81 & 14.66 & 42.87 & 5.41 & 69.76 & 86.33 & 43.17 \\
DeepSeek-V3.2-Chat & 53.43 & 14.60 & 28.24 & 78.65 & 56.65 & 12.27 & 14.56 & 57.87 & 5.72 & 5.72 & 86.60 & 37.66 \\
DeepSeek-V3.2-Reason & 57.22 & 9.72 & 25.88 & 81.28 & 60.67 & 13.89 & 10.91 & 57.07 & 7.27 & 66.92 & 87.08 & 43.45 \\

\midrule
\multicolumn{13}{c}{\cellcolor[HTML]{EFEFEF}\textit{\textbf{Closed-source LLMs}}} \\ \midrule
GPT-4o-mini & 59.26 & 11.21 & 37.57 & 72.04 & 58.59 & 11.63 & 17.54 & 62.46 & 6.52 & 89.32 & 80.82 & 46.09 \\
GPT5-mini & 57.78 & 9.74 & 29.16 & 87.33 & 60.33 & 11.45 & 13.63 & 66.76 & 4.28 & 55.15 & 87.00 & 43.87 \\
Qwen-turbo & 62.90 & 28.63 & 31.29 & 71.01 & 51.03 & 9.84 & 27.17 & 55.87 & 6.09 & 86.92 & 84.24 & 46.82 \\

\midrule
\end{tabular}}
\label{DeepSeek3.2_examiner_english}
\end{table*}

\begin{table*}[ht]
\centering
\caption{Evaluation results of baseline models on Chinese data assessed by Qwen3-Next-80B-A3B as the examiner model.}
\resizebox{\textwidth}{!}{
\begin{tabular}{
p{3.8cm}
>{\centering}p{1.3cm}
>{\centering}p{1.5cm}
>{\centering}p{1.2cm}
>{\centering}p{1.7cm}
>{\centering}p{1.7cm}
>{\centering}p{1.7cm}
>{\centering}p{1.7cm}
>{\centering}p{1.65cm}
>{\centering}p{1.7cm}
>{\centering}p{1.0cm}
>{\centering}p{1.6cm}
<{\centering}p{0.9cm}
<{\centering}}
\toprule
\textbf{Model} & \textbf{T\_Acc} & \textbf{E\_Recall} & \textbf{PD\_F1} & \textbf{PB\_Score} & \textbf{DD\_Score} & \textbf{TP\_IoU} & \textbf{CE\_Recall} & \textbf{CA\_IoU} & \textbf{CT\_IoU} & \textbf{FD\_F1} & \textbf{FB\_Score} & \textbf{Avg}   \\ \midrule

\multicolumn{13}{c}{\cellcolor[HTML]{EFEFEF}\textit{\textbf{Medical LLMs}}} \\ \midrule
Apollo2-7B & 56.63 & 31.63 & 34.59 & 70.98 & 42.84 & 7.14 & 16.05 & 59.36 & 2.21 & 70.16 & 76.66 & 42.57 \\
HuatuoGPT2-7B & 39.92 & 26.43 & 5.70 & 60.89 & 41.07 & 4.83 & 12.68 & 57.02 & 2.05 & 21.27 & 70.09 & 31.09 \\
HuatuoGPT2-13B & 61.57 & 33.64 & 0.00 & 62.42 & 38.60 & 6.47 & 17.27 & 60.36 & 2.24 & 3.86 & 70.81 & 32.48 \\
HuatuoGPT2-34B & 46.98 & 30.30 & 31.07 & 69.41 & 36.75 & 7.07 & 16.53 & 63.24 & 3.03 & 65.36 & 77.84 & 40.69 \\
MedGemma & 65.33 & 13.28 & 29.93 & 77.15 & 47.34 & 6.66 & 9.95 & 79.29 & 2.88 & 58.56 & 83.65 & 43.09 \\
HuatuoGPT-o1-7B & 67.14 & 27.15 & 32.25 & 71.18 & 38.95 & 5.35 & 11.32 & 68.78 & 2.76 & 60.99 & 77.05 & 42.08 \\
Baichuan-M2 & 53.65 & 25.74 & 33.94 & 78.87 & 46.65 & 5.32 & 13.26 & 77.71 & 2.60 & 65.97 & 81.40 & 44.10 \\

\midrule
\multicolumn{13}{c}{\cellcolor[HTML]{EFEFEF}\textit{\textbf{Open-source LLMs}}} \\ \midrule
Llama-3.3-70B & 64.94 & 17.25 & 29.61 & 73.55 & 37.30 & 5.59 & 9.42 & 70.52 & 2.33 & 68.27 & 84.85 & 42.15 \\
Llama-3.2-3B & 40.39 & 9.54 & 12.56 & 55.84 & 35.29 & 3.84 & 6.86 & 46.11 & 1.45 & 53.61 & 59.84 & 29.58 \\
Mistral-7B & 40.31 & 18.05 & 21.87 & 61.69 & 30.78 & 3.80 & 9.78 & 54.16 & 2.05 & 63.50 & 64.64 & 33.69 \\
Mixtral-8x22B & 62.90 & 28.49 & 27.37 & 70.12 & 31.55 & 5.97 & 13.99 & 70.67 & 2.43 & 67.39 & 80.22 & 41.92 \\
Falcon3-7B & 52.16 & 14.03 & 19.36 & 60.93 & 40.05 & 3.09 & 7.28 & 51.61 & 1.86 & 63.46 & 70.18 & 34.91 \\
Qwen2.5-72B & 65.02 & 31.51 & 36.43 & 78.56 & 53.41 & 8.58 & 14.99 & 76.17 & 2.91 & 67.67 & 83.75 & 47.18 \\
Qwen2.5-32B & 59.14 & 34.66 & 36.69 & 79.06 & 45.54 & 7.08 & 12.43 & 74.76 & 3.74 & 71.09 & 86.51 & 46.43 \\
Qwen2.5-14B & 64.16 & 24.17 & 37.60 & 78.76 & 41.62 & 6.57 & 11.48 & 74.18 & 2.86 & 69.49 & 80.53 & 44.67 \\
Qwen2.5-7B & 64.39 & 36.32 & 33.59 & 70.84 & 39.36 & 6.65 & 16.53 & 67.05 & 2.57 & 71.05 & 72.88 & 43.75 \\
Qwen3-Next-80B-A3B & 67.69 & 22.91 & 31.28 & 81.40 & 60.56 & 6.24 & 9.31 & 83.68 & 2.93 & 63.56 & 87.33 & 46.99 \\
DeepSeek-v3.2-Chat & 63.06 & 14.64 & 29.82 & 75.83 & 61.71 & 7.66 & 6.72 & 79.59 & 3.38 & 53.45 & 81.29 & 43.38 \\
DeepSeek-v3.2-Reason & 66.90 & 16.00 & 31.87 & 80.82 & 66.75 & 8.07 & 6.62 & 84.45 & 3.54 & 56.49 & 86.46 & 46.18 \\

\midrule
\multicolumn{13}{c}{\cellcolor[HTML]{EFEFEF}\textit{\textbf{Closed-source LLMs}}} \\ \midrule
GPT-4o-mini & 63.38 & 16.98 & 28.55 & 71.46 & 40.56 & 5.75 & 8.40 & 71.26 & 2.51 & 71.72 & 83.76 & 42.21 \\
GPT-5-mini & 61.97 & 3.21 & 11.59 & 83.19 & 50.89 & 6.12 & 1.01 & 83.77 & 3.90 & 44.01 & 82.35 & 39.27 \\
Qwen-turbo & 53.99 & 31.63 & 26.61 & 79.72 & 53.52 & 6.10 & 7.78 & 73.77 & 3.16 & 74.86 & 83.10 & 44.93 \\

\midrule
\end{tabular}}
\label{qwen3_examiner_chinese}
\end{table*}

\begin{table*}[ht]
\centering
\caption{Evaluation results of baseline models on Chinese data assessed by DeepSeek-V3.2 as the examiner model.}
\resizebox{\textwidth}{!}{
\begin{tabular}{
p{3.8cm}
>{\centering}p{1.3cm}
>{\centering}p{1.5cm}
>{\centering}p{1.2cm}
>{\centering}p{1.7cm}
>{\centering}p{1.7cm}
>{\centering}p{1.7cm}
>{\centering}p{1.7cm}
>{\centering}p{1.65cm}
>{\centering}p{1.7cm}
>{\centering}p{1.0cm}
>{\centering}p{1.6cm}
<{\centering}p{0.9cm}
<{\centering}}
\toprule
\textbf{Model} & \textbf{T\_Acc} & \textbf{E\_Recall} & \textbf{PD\_F1} & \textbf{PB\_Score} & \textbf{DD\_Score} & \textbf{TP\_IoU} & \textbf{CE\_Recall} & \textbf{CA\_IoU} & \textbf{CT\_IoU} & \textbf{FD\_F1} & \textbf{FB\_Score} & \textbf{Avg}   \\ \midrule

\multicolumn{13}{c}{\cellcolor[HTML]{EFEFEF}\textit{\textbf{Medical LLMs}}} \\ \midrule
Apollo2-7B & 57.74 & 33.57 & 34.59 & 70.88 & 43.29 & 7.45 & 18.25 & 59.65 & 2.31 & 70.31 & 76.58 & 43.15 \\
HuatuoGPT2-7B & 40.63 & 29.06 & 5.43 & 60.94 & 41.52 & 5.00 & 19.50 & 56.74 & 2.24 & 22.42 & 70.93 & 32.22 \\
HuatuoGPT2-13B & 61.33 & 35.65 & 0.00 & 62.59 & 38.26 & 6.55 & 21.14 & 60.15 & 2.20 & 3.70 & 70.02 & 32.87 \\
HuatuoGPT2-34B & 45.18 & 29.60 & 29.42 & 66.49 & 35.14 & 6.30 & 18.42 & 60.74 & 2.75 & 62.36 & 75.37 & 39.25 \\
MedGemma & 65.49 & 16.49 & 30.15 & 77.22 & 47.67 & 7.08 & 11.02 & 79.10 & 2.95 & 58.71 & 83.87 & 43.61 \\
HuatuoGPT-o1-7B & 66.67 & 22.62 & 32.91 & 70.53 & 39.22 & 5.61 & 14.49 & 68.52 & 2.57 & 62.25 & 76.88 & 42.02 \\
Baichuan-M2 & 53.25 & 26.06 & 35.54 & 78.82 & 46.98 & 5.29 & 14.99 & 76.47 & 2.49 & 65.20 & 82.16 & 44.30 \\

\midrule
\multicolumn{13}{c}{\cellcolor[HTML]{EFEFEF}\textit{\textbf{Open-source LLMs}}} \\ \midrule
Llama-3.3-70B & 65.02 & 18.44 & 29.43 & 72.93 & 37.76 & 5.53 & 12.93 & 69.98 & 2.35 & 67.70 & 84.67 & 42.43 \\
Llama-3.2-3B & 41.41 & 14.19 & 12.45 & 55.91 & 34.68 & 4.05 & 8.21 & 45.78 & 1.54 & 53.10 & 60.64 & 30.18 \\
Mistral-7B & 40.63 & 18.62 & 22.09 & 61.68 & 30.23 & 3.75 & 10.95 & 53.93 & 1.89 & 63.68 & 64.06 & 33.77 \\
Mixtral-8x22B & 62.67 & 30.21 & 27.56 & 70.85 & 32.03 & 5.91 & 15.17 & 70.93 & 2.38 & 66.84 & 80.71 & 42.30 \\
Falcon3-7B & 51.84 & 19.12 & 19.52 & 60.11 & 39.39 & 3.19 & 15.81 & 52.93 & 1.86 & 62.52 & 70.46 & 36.07 \\
Qwen2.5-72B & 65.10 & 27.79 & 36.78 & 78.37 & 53.95 & 8.36 & 15.77 & 76.56 & 2.89 & 67.79 & 83.59 & 47.00 \\
Qwen2.5-32B & 59.06 & 31.12 & 36.78 & 79.07 & 44.74 & 7.31 & 13.05 & 74.60 & 3.53 & 70.87 & 86.57 & 46.06 \\
Qwen2.5-14B & 63.92 & 23.20 & 37.32 & 78.13 & 41.71 & 6.76 & 11.76 & 73.53 & 2.92 & 69.14 & 80.11 & 44.41 \\
Qwen2.5-7B & 64.39 & 31.76 & 33.42 & 70.56 & 38.92 & 6.74 & 16.78 & 67.41 & 2.46 & 71.05 & 73.11 & 43.33 \\
Qwen3-Next-80B-A3B & 68.08 & 21.97 & 31.57 & 82.10 & 60.80 & 6.12 & 10.13 & 84.74 & 2.94 & 63.74 & 88.30 & 47.32 \\
DeepSeek-v3.2-Chat & 66.20 & 21.46 & 31.35 & 80.55 & 65.63 & 8.00 & 11.95 & 85.18 & 3.46 & 56.75 & 86.93 & 47.04 \\
DeepSeek-v3.2-Reason & 66.98 & 21.58 & 32.06 & 80.85 & 65.32 & 7.95 & 11.62 & 85.14 & 3.68 & 56.62 & 86.10 & 47.08 \\

\midrule
\multicolumn{13}{c}{\cellcolor[HTML]{EFEFEF}\textit{\textbf{Closed-source LLMs}}} \\ \midrule
GPT-4o-mini & 63.38 & 19.52 & 28.88 & 71.55 & 40.00 & 5.75 & 10.95 & 71.21 & 2.07 & 71.93 & 84.32 & 42.69 \\
GPT-5-mini & 58.22 & 4.81 & 12.70 & 83.19 & 50.42 & 6.30 & 5.87 & 83.49 & 3.90 & 44.73 & 83.00 & 39.69 \\
Qwen-turbo & 52.58 & 30.76 & 27.67 & 78.03 & 52.86 & 5.84 & 9.72 & 72.17 & 2.83 & 75.30 & 83.47 & 44.66 \\

\midrule
\end{tabular}}
\label{DeepSeek3.2_examiner_chinese}
\end{table*}

\subsection{Error Case}
\label{error_case}


In this section, we introduce the Chinese and English error samples of LLMs on ClinicalMC. Both error examples come from the DeepSeek-V3 model.
The Chinese error sample is shown in Fig.~\ref{zh_error_case}. 
The English error sample is shown in Fig.~\ref{en_error_case}.

\definecolor{c1}{HTML}{2BB45C}
\definecolor{c2}{HTML}{F84C4D}
\definecolor{c3}{HTML}{BA9325}

\begin{figure*}[t]
\centering
\includegraphics[width=\textwidth]{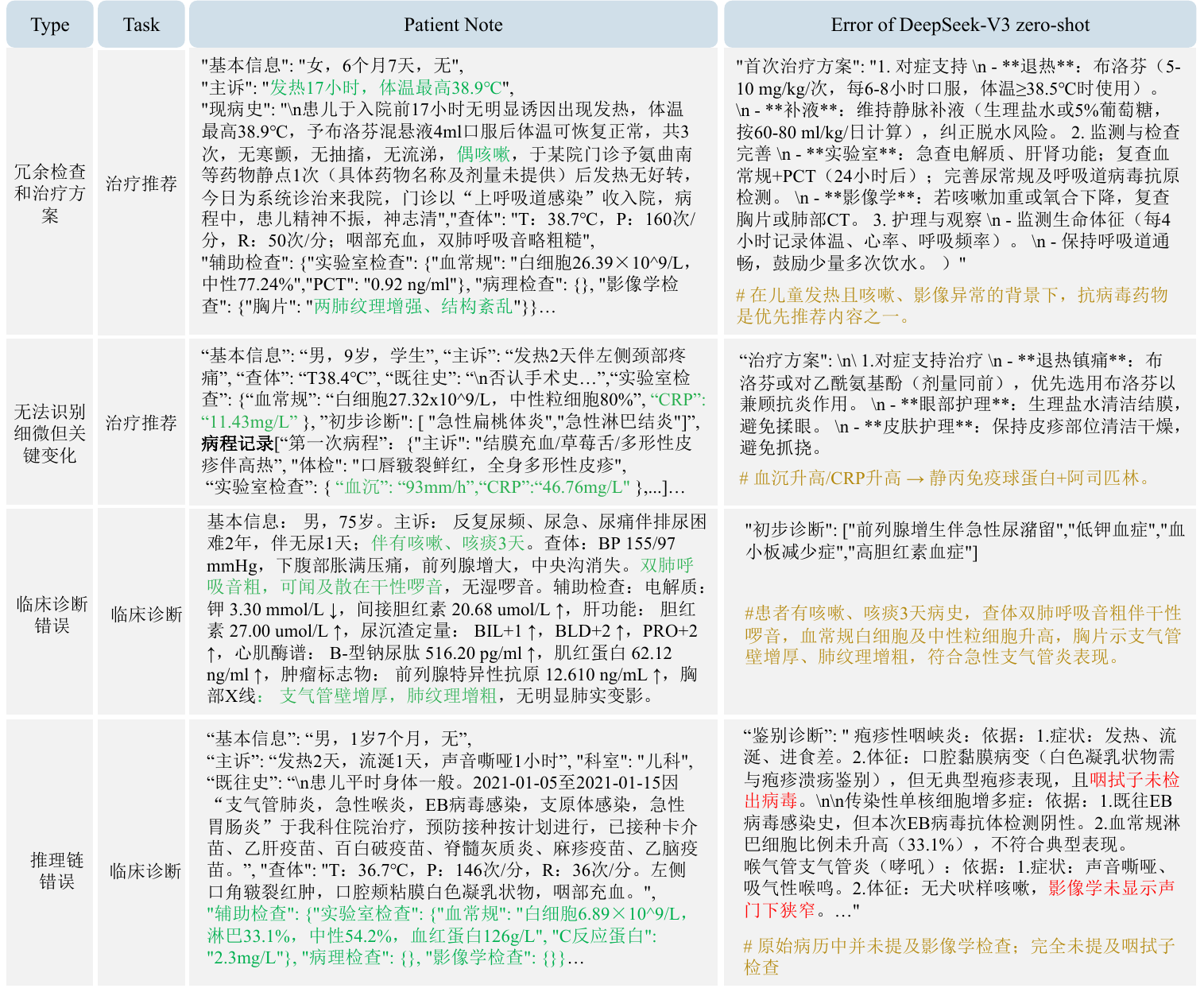} 
\caption{Examples of the three error types for Chinese data in ClinicalMC. The \textcolor{c2}{incorrect rationale}, \textcolor{c3}{\# comments}, and \textcolor{c1}{evidence} are highlighted. }
\label{zh_error_case}
\end{figure*}

\begin{figure*}[t]
\centering
\includegraphics[width=\textwidth]{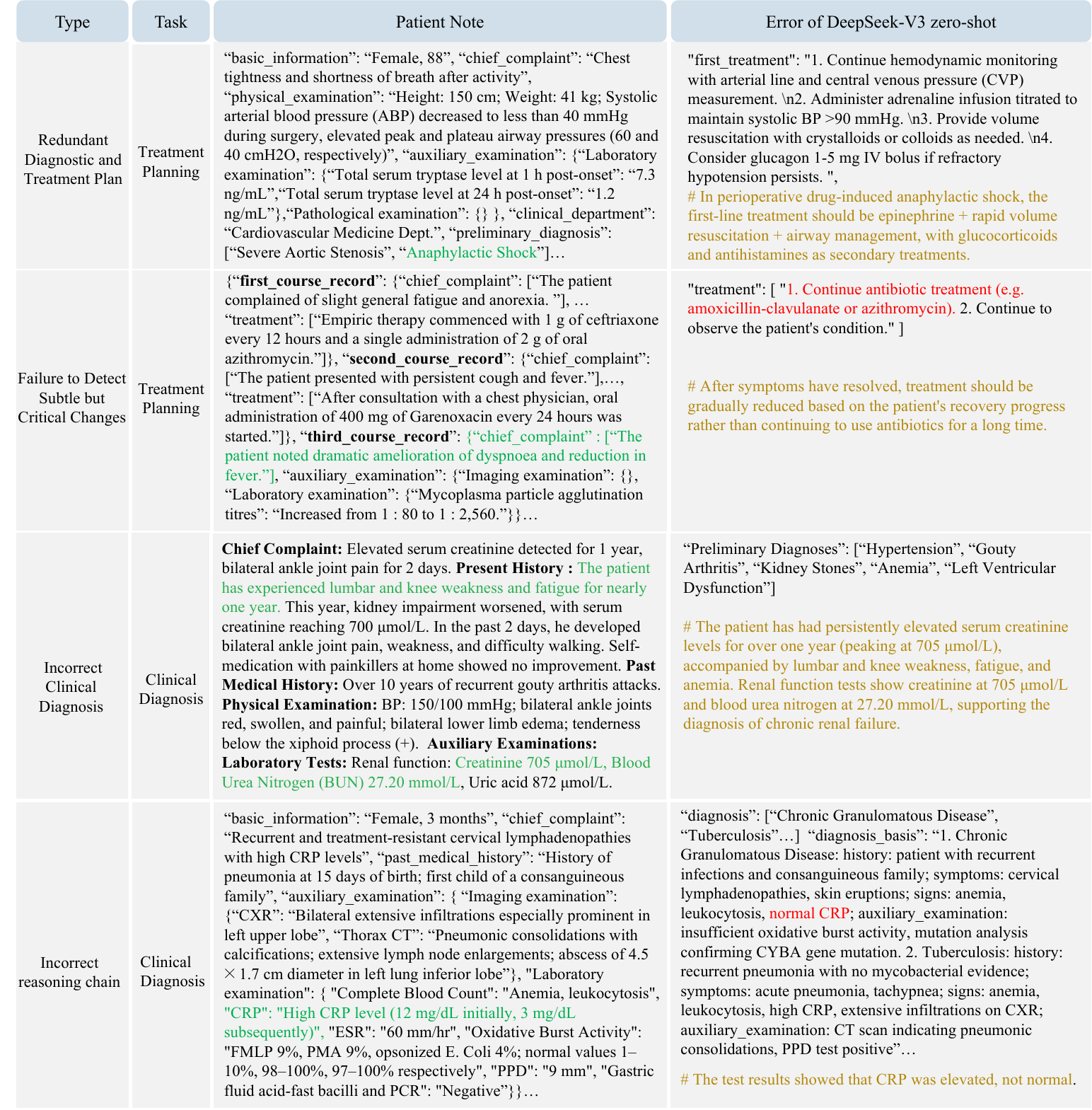} 
\caption{Examples of the three error types for English data in ClinicalMC. The \textcolor{c2}{incorrect rationale}, \textcolor{c3}{\# comments}, and \textcolor{c1}{evidence} are highlighted. }
\label{en_error_case}
\end{figure*}

\subsection{Prompt of SimHospital Framework}
\label{framework_prompt}

\begin{figure*}[h]
\centering
\includegraphics[width=\textwidth]{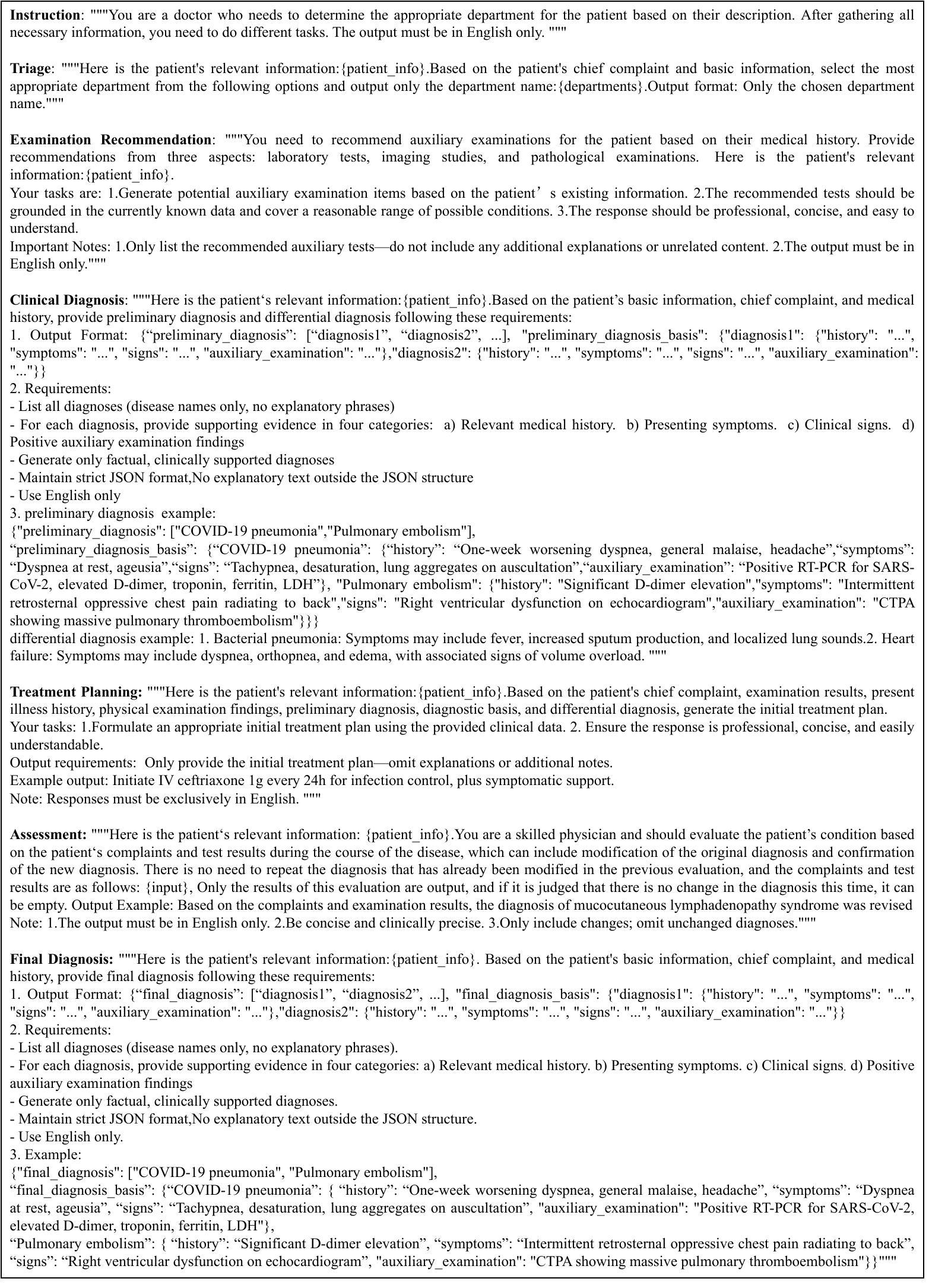} 
\caption{Prompt of the doctor agent.}
\label{doctor_prompt1}
\end{figure*}

\begin{figure*}[h]
\centering
\includegraphics[width=\textwidth]{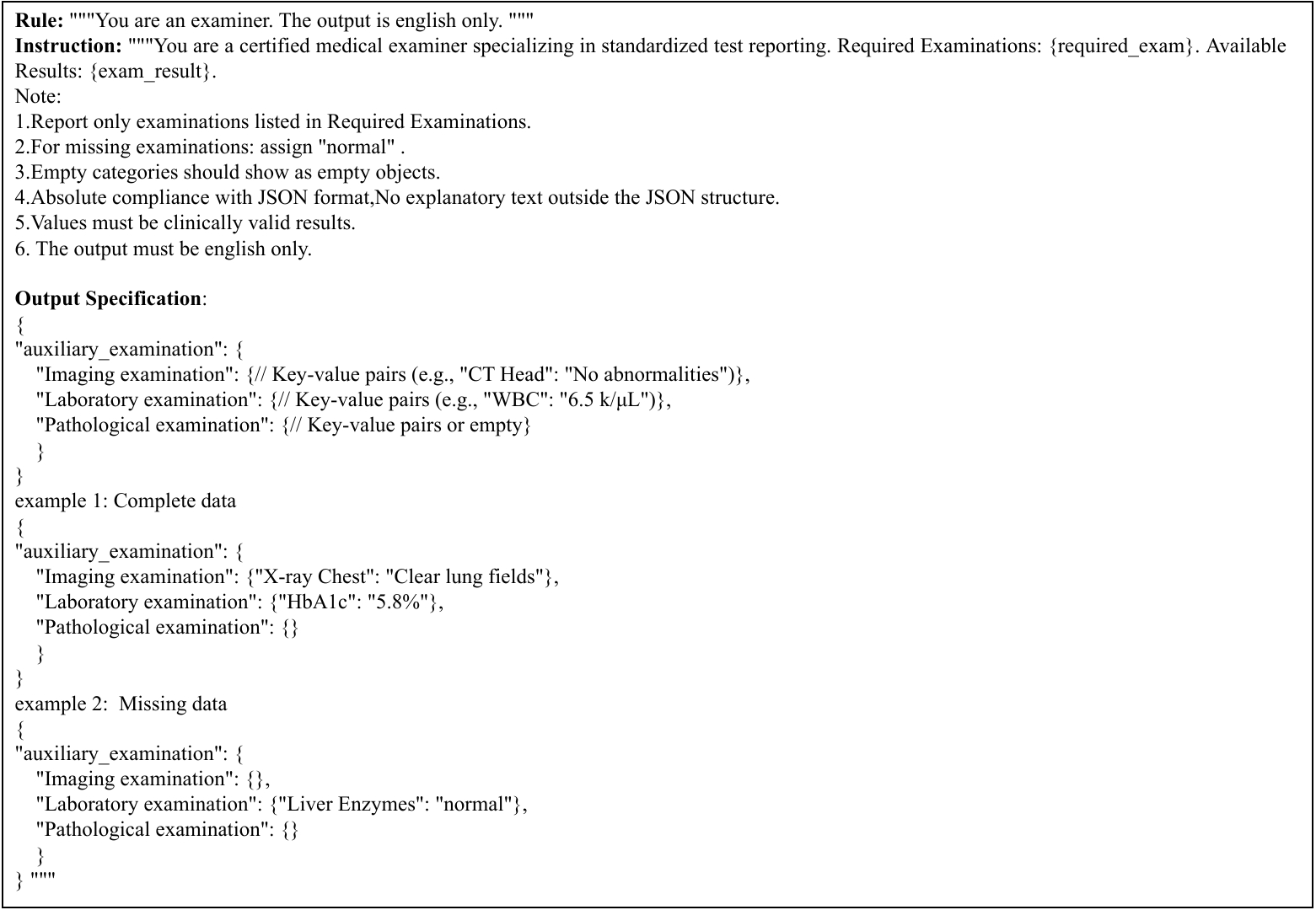} 
\caption{Prompt of the examiner agent.}
\label{examiner_prompt}
\end{figure*}

\begin{figure*}[h]
\centering
\includegraphics[width=\textwidth]{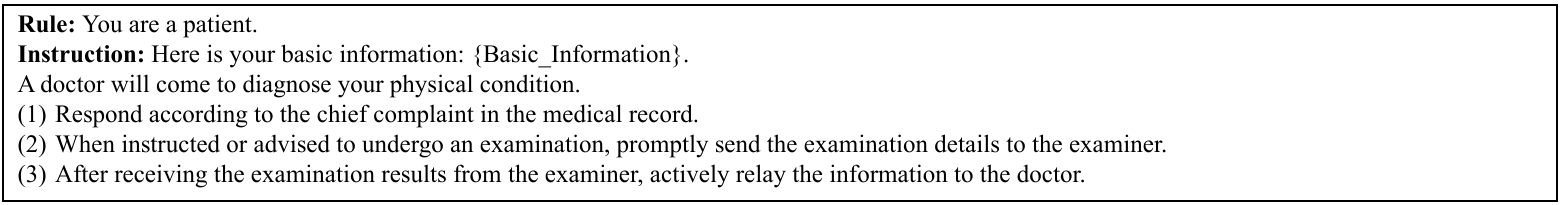} 
\caption{Prompt of the patient agent.}
\label{patient_prompt}
\end{figure*}

In this section, we provide a detailed description of the prompts for the three agents introduced in the SimHospital evaluation framework. 
The prompt for the doctor agent is shown in Fig.~\ref{doctor_prompt1}. The prompt for the examiner agent is shown in Fig.~\ref{examiner_prompt}. The prompt for the patient agent is shown in Fig.~\ref{patient_prompt}.


\subsection{Prompt of ClinicalMC Annotation}
\label{annotation_prompt}

\begin{figure*}[h]
\centering
\includegraphics[width=\textwidth]{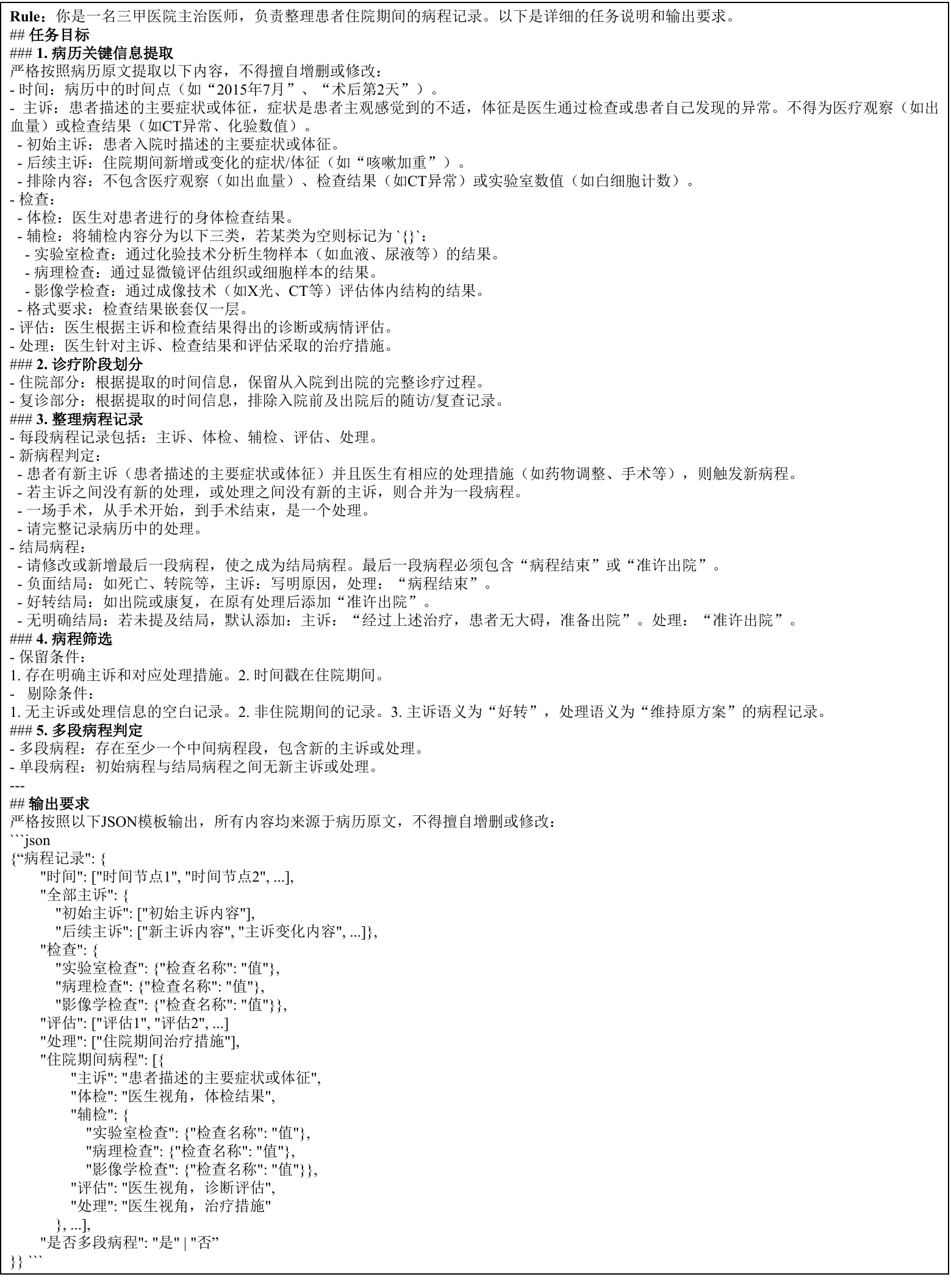} 
\caption{Prompt for data annotation.}
\label{trans_prompt}
\end{figure*}

In this section, we provide a detailed description of the prompts used during the ClinicalMC annotation process. During data annotation, the model is explicitly instructed to ``strictly extract the following information from the original medical records without adding, deleting, or modifying any content.'' to minimize hallucination during the annotation. 
The prompt for data annotation is shown in  Fig.~\ref{trans_prompt}.

\subsection{Evaluation Prompts and ClinicalMC Examples}
\label{example}

\begin{figure*}[h]
\centering
\includegraphics[width=\textwidth]{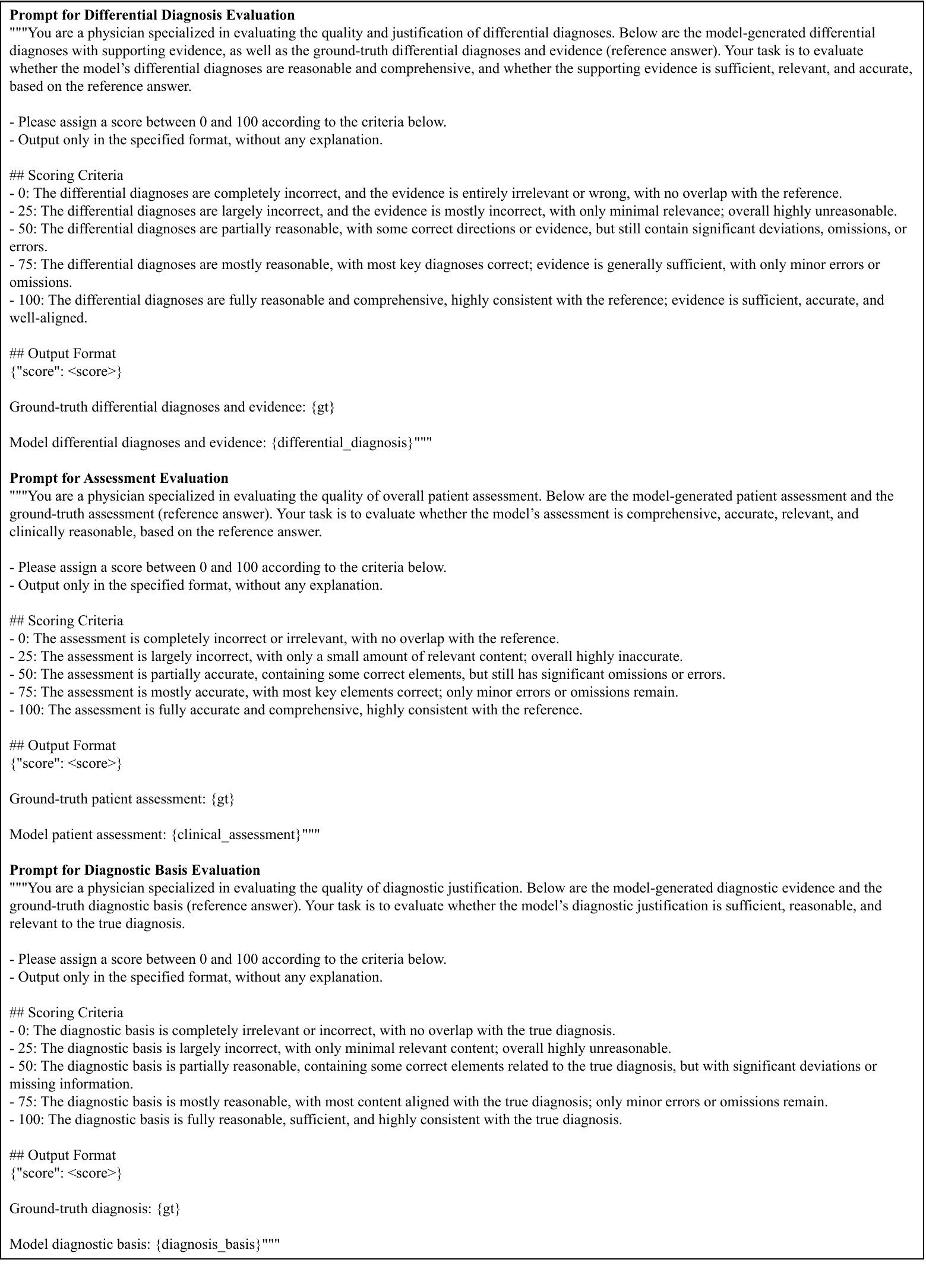} 
\caption{Prompts for evaluating differential diagnosis, diagnostic basis, and assessment.}
\label{evaluation_prompt}
\end{figure*}
\begin{figure*}[h]
\centering
\includegraphics[width=\textwidth]{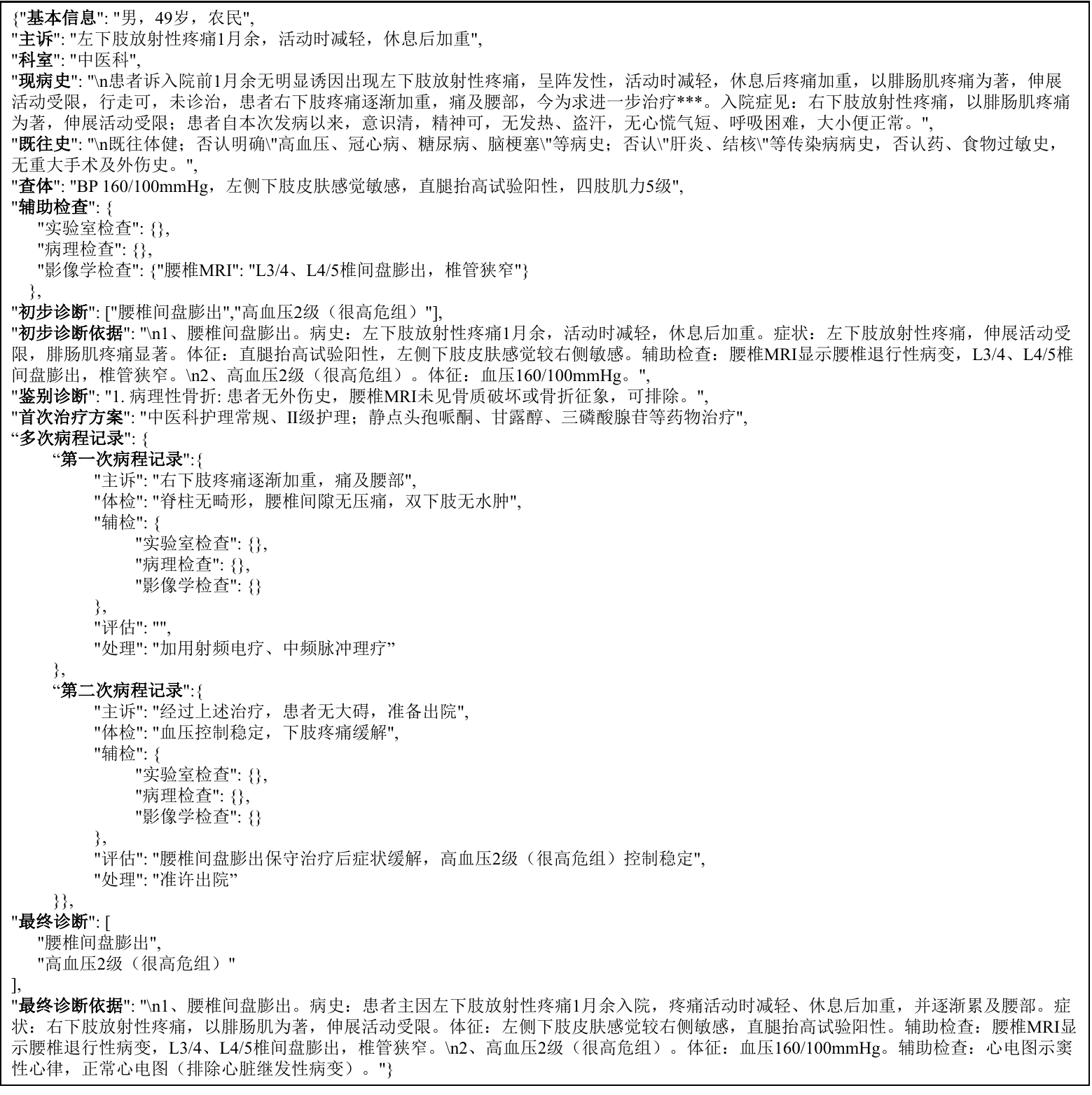} 
\caption{Chinese EHR example.}
\label{full_example2}
\end{figure*}
\begin{figure*}[h]
\centering
\includegraphics[width=\textwidth]{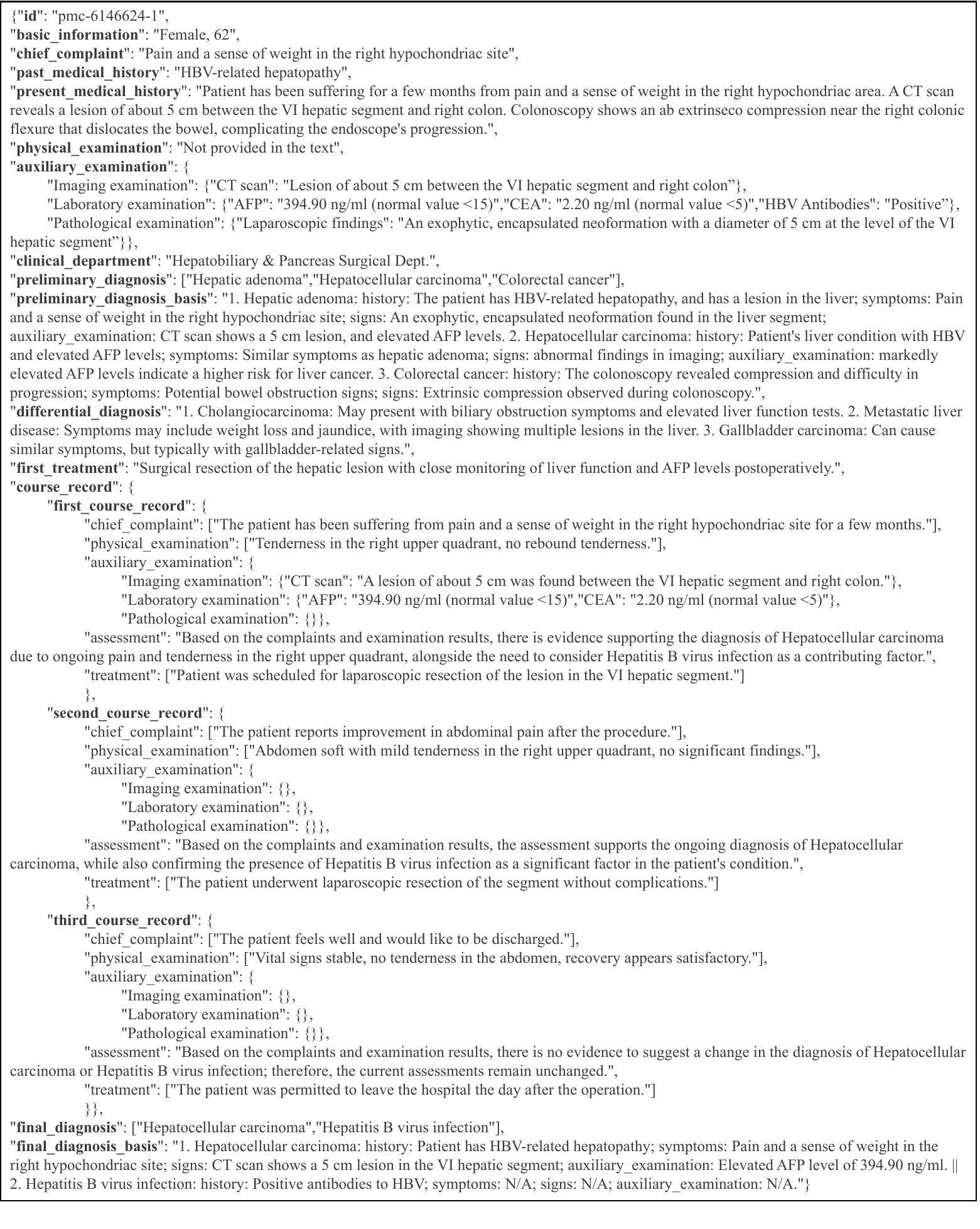} 
\caption{English EHR example.}
\label{full_example}
\end{figure*}

In this section, we present the evaluation prompts as well as example Chinese and English EHRs from ClinicalMC. 
The prompts used for evaluation are shown in Fig.~\ref{evaluation_prompt}. 
The Chinese EHR is shown in Fig.~\ref{full_example2}, and the English EHR is shown in Fig.~\ref{full_example}.

\end{document}